\documentclass[twoside]{article}

\usepackage{amsmath,amsfonts,bm}

\def\eqref#1{equation~\ref{#1}}

\def\1{\bm{1}}

\DeclareMathAlphabet{\mathsfit}{\encodingdefault}{\sfdefault}{m}{sl}
\SetMathAlphabet{\mathsfit}{bold}{\encodingdefault}{\sfdefault}{bx}{n}

\usepackage[accepted]{aistats2026}

\usepackage{hyperref}
\usepackage{url}
\usepackage{enumitem}
\usepackage{amstext}
\usepackage{amsthm}

\usepackage{amsmath}
\usepackage{xcolor}
\usepackage{comment}
\usepackage{tikz}
\usetikzlibrary{decorations.pathmorphing}
\usepackage{multirow}
\usepackage{booktabs} 
\usepackage{caption} 
\usepackage{graphicx} 
\usepackage[linesnumbered,ruled,vlined]{algorithm2e}

\newtheorem{proposition}{Proposition}
\usepackage{stfloats}
\usepackage{subcaption} 
\newcommand{\vect}[1]{\boldsymbol{#1}}

\DeclareMathOperator*{\divergence}{div}

\definecolor{darkgreen}{HTML}{2E7D32}
\definecolor{salmon}{rgb}{0.98,0.50,0.45}
\definecolor{grey}{gray}{0.6}
\usepackage{pifont}

\graphicspath{{../}}

\begin{document}

%

%

\twocolumn[
\aistatstitle{Variational Grey-Box Dynamics Matching}
\aistatsauthor{Gurjeet Sangra Singh \And  Frantzeska Lavda \And  Giangiacomo Mercatali \And Alexandros Kalousis}
\aistatsaddress{University of Geneva \\ HES-SO Geneva \And HES-SO Geneva \And HES-SO Geneva \And HES-SO Geneva}
]

\begin{abstract}
Deep generative models such as flow matching and diffusion models have shown great potential in learning complex distributions and dynamical systems, but often act as black-boxes, neglecting underlying physics. In contrast, physics-based simulation models described by ODEs/PDEs remain interpretable, but may have missing or unknown terms, unable to fully describe real-world observations.
We bridge this gap with a novel grey-box method that integrates incomplete physics models directly into generative models. Our approach learns dynamics from observational trajectories alone, without ground-truth physics parameters, in a simulation-free manner that avoids scalability and stability issues of Neural ODEs. The core of our method lies in modelling a structured variational distribution within the flow matching framework, by using two latent encodings: one to model the missing stochasticity and multi-modal velocity, and a second to encode physics parameters as a latent variable with a physics-informed prior. Furthermore, we present an adaptation of the framework to handle second-order dynamics.
Our experiments on representative ODE/PDE problems and real-world weather forecasting demonstrate that our method performs on par with or superior to fully data-driven approaches and previous grey-box baselines, while preserving the interpretability of the physics model.
Our code is available at \url{https://github.com/DMML-Geneva/VGB-DM}.
\end{abstract}

\section{Introduction}

Modelling the dynamics of physical systems is a fundamental challenge in various scientific domains, e.g., in fluid dynamics, geophysics and others. Such systems are described by physics models, often formulated as Ordinary or Partial Differential Equations (ODEs/PDEs). Because they are derived from physical laws such as conservation of energy, momentum, or mass, these models provide interpretable parameters and equations whose terms have clear physical meaning. 
However, real-world systems often exceed the complexity captured by known physics, resulting in incomplete models with unknown parameters or missing terms.  

Deep generative models, by contrast, have shown remarkable success in learning complex data distributions. Techniques such as Neural ODEs \cite{chen2018neuralode} or Flow Matching \cite{lipman2023, albergo2023,liu2023flow} can model high-dimensional distributions effectively. Applied to dynamical systems, however, these methods act as black boxes; they learn the dynamics purely from observations without leveraging existing physical knowledge. This can lead to physically implausible predictions, poor generalization,
and limited interpretability.

To bridge this gap, deep grey-box modelling approaches aim to integrate incomplete physics knowledge with data-driven methods.
Different simulation-based methods, which leverage Neural ODEs \cite{chen2018neuralode}, such as Physics-Integrated Variational Autoencoders (PhysVAE) \cite{naoya_2021} and related methods \cite{yin_2021, Mehta2020NeuralDS, qian2021pharma_ode, wehenkel2023robust, verma2024climode} combine physics models with neural networks. While effective, these methods rely heavily on numerical ODE solvers during training, which can introduce scalability issues, numerical instability, and significant computational overhead.

Furthermore, learning the dynamics of physical systems often involves dealing with inherent stochasticity or multi-modal velocity fields. When physics is incomplete or parameters are unknown, multiple valid future trajectories might exist from the same state. Standard deterministic approaches struggle to capture this complexity. Recent advances in Variational Rectified Flow Matching \cite{guo2025variationalrectifiedflowmatching} highlight the importance of introducing stochasticity via latent variables to learn more expressive and accurate vector fields, arguing that this is essential for capturing complex dynamics.

In this work, we introduce Variational Grey-Box Dynamics Matching (VGB-DM)\footnote{We have open-sourced our VGB-DM model and experiments in our \href{https://github.com/DMML-Geneva/VGB-DM}{VGB-DM GitHub repository}}, a novel approach that integrates incomplete physics models into a simulation-free generative modelling paradigm while explicitly accounting for stochasticity and parameter uncertainty.  Our approach builds upon the simulation-free methods, adapting them to learn dynamics directly from trajectories \cite{zhang2024tjfm}. We employ a variational approach inspired by \cite{guo2025variationalrectifiedflowmatching}, but introduce a structural latent space tailored for grey-box modelling, one part represents random variability in the system’s dynamics, while another part represents physical parameters that must be inferred from data.

Our contributions are summarized as follows:
\begin{enumerate}
    \item Simulation-Free Grey-Box modelling: We propose a novel integration of incomplete physics models in a generative model inspired by gradient matching as used in recent Flow Matching generative models. This approach learns dynamics and infers parameters in a simulation-free manner, avoiding the computational bottlenecks, memory scalability and stability issues associated with solver-based methods like Neural ODEs, enhancing scalability for high-dimensional grey-box modelling.
    \item Structured Variational Inference: We introduce a structured variational distribution over the velocity field, utilising separate latent variables for modelling stochasticity/multi-modality and inferring physics parameters. This allows the model to capture complex dynamics while maintaining physics interpretability.
    \item Empirical Validation: We demonstrate the effectiveness of VGB-DM on different ODE/PDE systems and weather forecasting task, showing superior performance and faster convergence compared to state-of-the-art baselines.
\end{enumerate}

\section{Background} \label{sec:background}
\subsection{Problem setting} \label{sec:problem_settings}
Consider an unknown data-generating process (DGP) underlying the real-world dynamical system. We observe data as a collection of trajectories $\mathcal{D}=\{\vect{x}^i\}_{i=1}^N$. Each sample $\vect{x}^i$ corresponds to a time series, $\vect{x}^i = [x^i_0, \dots, x^i_T]$, where $x^i_t \in \mathcal{X} \subseteq \mathbb{R}^d$ is the state at any discretized time $t \in \{0,\dots, T\}$. These trajectories are assumed to be independent and identically distributed (i.i.d.) samples from an unknown \textit{data} distribution $\pi(\vect{x}) \in \mathcal{P}(\mathcal{X}^{T+1})$, where $\mathcal{P}(\mathcal{X}^{T+1})$ is a set of all probability measures over the trajectory space $\mathcal{X}^{T+1}$.

A central goal in learning dynamical systems with generative models is to construct a conditional model that, given a current state $x_{t}$ and possibly an arbitrary history of past observations $\vect{x}_{t-h:t}$ with $h \in \mathbb{Z}^+$, can generate plausible future trajectories. Such a model should forecast consistently by learning the underlying system evolution, including extrapolation beyond observed horizons, resulting into approximating the true conditional distribution $\pi(\vect{x} \mid \vect{x}_{t-h:t})$.

\subsection{Grey-Box Modelling: Beyond Forecasting}
Within the problem description given earlier, grey-box modelling approaches have demonstrated capabilities in generating robust dynamics and physics parameter inference \cite{naoya_2021, wehenkel2023robust}. In contrast to purely data-driven forecasting, the grey-box setting assumes the availability of an approximate but useful physics model. This introduces an additional goal: not only generating consistent system dynamics, but also inferring underlying physics parameters from observed trajectories. Making them  well-suited for real-world scenarios where physics knowledge is incomplete but valuable. In this framework, the physics model typically comes in the form of ordinary or partial differential equations (ODEs/PDEs):
$\frac{\partial x_t}{\partial t} = f_p(x_t, \theta)$,
where $f_p: \mathcal{X}\times \Theta \mapsto \mathbb{R}^d$ is the known (but potentially incomplete) physics model and $\theta \in \Theta \subseteq \mathbb{R}^p$ represents some interpretable physics parameters, often inferred.
Deep grey-box models learn the time evolution of the dynamic system by leveraging some physics model by augmenting it with some Deep Neural Network (DNN):
$\frac{\partial x_t}{\partial t} = f_\phi (x_t)\circ f_p(x_t, \theta)$,
where $f_\phi: \mathcal{X} \mapsto \mathbb{R}^d$ is a DNN learned from data, and the operand $\circ$ the function composition operator (even if the additive form, $f_\phi(x_t) + f_p(x_t, \theta)$, is prevalent in the literature, we adopt a more general notation here). For example, in an RLC circuit, the dynamics of the grey-box can be described by:
\begin{align*} \frac{\partial}{\partial t} \begin{bmatrix} I_t \ U_t  \end{bmatrix} = \underbrace{\begin{bmatrix} \frac{I_t}{C} \ \frac{1}{L}(V_t - U_t) \end{bmatrix}}_{f_p} + \underbrace{\begin{bmatrix} 0 \ -\frac{R}{L}I_t \end{bmatrix}}_{\substack{f_\phi}}, 
\end{align*} where $f_p$ represents the known capacitor and inductor dynamics with $\theta=[L, C]$, and $f_\phi$ to be learned by the DNN, which captures the missing physics and the unknown resistance term.

However, a critical challenge arises from this formulation: in real-world systems, only state trajectories are observed, making the usage of the physics model more challenging, where physics' parameters are not paired. Estimating these parameters constitutes a challenging inverse problem due to potential ill-posedness and data limitations. To address this, generative grey-box models (e.g. \cite{naoya_2021, wehenkel2023robust, verma2024climode} solutions) embed inference mechanism, such as encoders that ground latent spaces in physical principles. This structure is illustrated in the probabilistic graphical model in Figure~\ref{fig:PGM}.
Therefore, the goal in grey-box modelling extends beyond forecasting dynamics to include inference of the underlying physics parameters from observed data, while validating their meaningful usage within the model.

\begin{figure}[t!]
    \centering
    \vspace{0.1cm}
    \resizebox{0.46\textwidth}{!}{ 
    \begin{tikzpicture}
        \node[circle, draw, fill=grey!30, minimum size=1cm] (init) at (2, 4) {$\vect{x}_0$};
        \node[circle, draw=blue, dash pattern=on 1pt off 1.5pt, line width=1pt, minimum size=1cm] (theta) at (2,2) {$\vect \theta$};
        \node[circle, draw=blue, dash pattern=on 1pt off 1.5pt, line width=1pt, minimum size=1cm] (z) at  (5,4) {$\vect z$};
        \node[circle, draw, fill=grey!30, minimum size=1cm] (x) at (5,2) {${\vect{x}}$};
        
        \draw[dashed, ->] (z) -- (x);
        \draw[dashed, ->] (theta) -- (x);
        \draw[->] (init) -- (x);
        
        \draw[black, thick, dashed] (0.3, 1.3) rectangle (4.0, 5.0);
        \draw[black, thick] (4.3, 1.3) rectangle (5.7, 5.0);
        
        \node[anchor=north west] at (0.3, 5.) {\small Physics model};
        
        
        \node[anchor=north west] at (6.0, 5.0) {\small \textbf{Legend:}};
        \node[circle, draw, fill=grey!30, minimum size=0.3cm] at (6.2, 4.2) {};
        \node[anchor=west] at (6.5, 4.2) {\small Observed};
        \node[circle, draw=blue, dash pattern=on 1pt off 1.5pt, line width=0.8pt, minimum size=0.3cm] at (6.2, 3.8) {};
        \node[anchor=west] at (6.5, 3.8) {\small Latent};
        \draw[->] (6.0, 3.4) -- (6.4, 3.4);
        \node[anchor=west] at (6.5, 3.4) {\small Deterministic};
        \draw[dashed, ->] (6.0, 3.) -- (6.4, 3.0);
        \node[anchor=west] at (6.5, 3.0) {\small Inferred};
    \end{tikzpicture}
    } 
    \caption{Probabilistic graphical model of variational grey-box models. Shaded circles are observed variables, empty dashed circles are latent variables, and dashed arrows indicate learned inference dependencies.}
    \label{fig:PGM}
\end{figure}

\subsection{Flow Matching}
Simulation-free generative models such as Flow Matching variant algorithms \cite{lipman2023, albergo2023, liu2023flow} have achieved great success in deep generative models. Given a known source distribution $\pi_0(x_0)$ (typically a standard Gaussian) and a target data distribution $\pi_1(x_1)$, flow matching algorithms \cite{lipman2023, albergo2023, liu2023flow} learn a time-dependent vector field $v_t^\phi : [0,1] \times \mathbb{R}^d \rightarrow \mathbb{R}^d$ that defines a continuous transformation of samples from $x_0 \sim \pi_0$ to $ x_1 \sim \pi_1$. This transformation is governed by the continuity equation:
\begin{align*}
    \frac{\partial p_t^\phi(x_t)}{\partial t} = - \divergence(p_t^\phi(x_t) v_t^\phi(x_t)),
\end{align*}
where $p_t$ is the marginal density at time $t \in [0,1]$ .
Instead of solving this equation through expensive simulations of forward dynamics, Conditional Flow Matching (CFM) learns directly $v_t^\phi$ by regressing it to a target velocity field by constructing a path of tractable conditional probabilities $p_t( x_t | x_1)$ defined along a linear interpolation from source samples to the data-target samples distribution, $x_t = (1 - t) x_0 + t x_1$, with $(x_0, x_1) \sim \pi(x_0, x_1)$ and $\dot{x}_t = \frac{d}{d t}x_t=x_1 - x_0$. The training loss is then:
\begin{align}
    \mathcal{L}_{\text{CFM}}(\phi) = \mathbb{E}_{\substack{\\ x_0 \sim \pi_0, x_1 \sim \pi_1 \\ t \sim \mathcal{U}(0,1)}}
    \left[ \left\| v_t^\phi(x_t) - \dot{x}_t \right\|^2 \right],
\end{align}
yielding a simulation-free objective. While linear interpolation is most common, Flow Matching algorithms also admit more general deterministic or stochastic paths, which influence the bias–variance tradeoffs of the estimator \cite{albergo2023, liu2023flow}.

\begin{figure*}[h]
\centering
\includegraphics[width=0.999\textwidth]{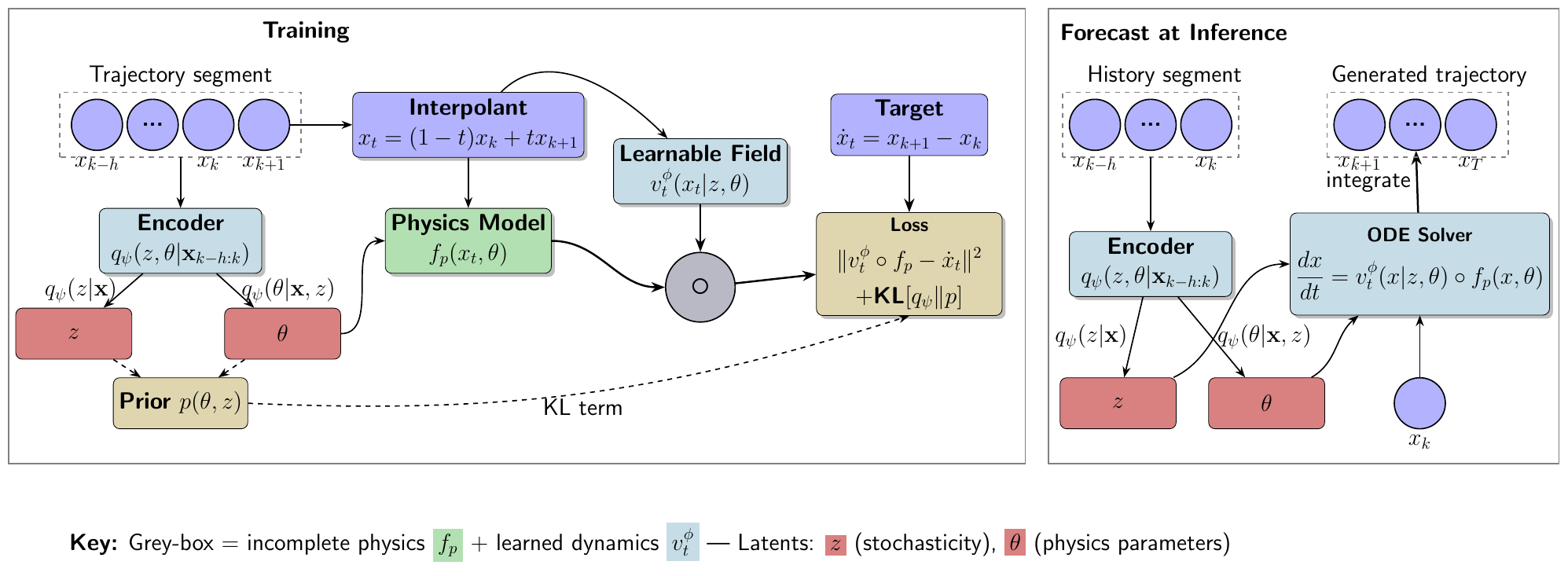}
\caption{Overview of the Variational Grey-Box Dynamics Matching (VGB-DM) framework.}
\label{fig:main-diagram}
\end{figure*}

\section{Method} \label{sec:method}
We now present multiple building block choices and modelling aspects of our algorithm, yielding to a generative model specifically designed for solving dynamical systems, which integrates physics priors in the form of physics differential equations fully trained in a simulation-free manner, and physics structured latent variables modelled by a variational distribution. The latter models physics latent variables and accounts for missing stochasticity. We denote this method as Variational Grey-Box Dynamic Matching (VGB-DM), as our goal is to learn dynamical systems by integrating physics-based components and grounding our generative model on the physics.

\subsection{Variational Grey-Box Dynamics Matching}

\paragraph{Integrating Incomplete Physics.}
Contrary to Conditional Flow Matching, which addresses generative tasks by mapping samples from noise to data samples, our setting focuses on dynamical systems where data naturally comes as time-series trajectories. We exploit the structure of the trajectory to optimize a target velocity vector field by building it on consecutive states of the trajectory, in the spirit of Trajectory Flow Matching (TFM) \cite{zhang2024tjfm}, but extend it to incorporate incomplete physics $f_p$, which encode the underlying dynamics of the system. 

More formally, let $\pi(\vect{x})$ denote the data distribution over trajectories $\vect{x}^{i} = (x_0^{i}, x_1^{i}, \dots, x_T^{i})$, and by $\vect x_{k-h:k+1} \sim \pi(\cdot | \vect{x})$ the distribution over $(x_{k-h},\dots, x_{k}, x_{k+1})$ pairs of consecutive points induced by the trajectory with $k \in \mathcal{U}\{h, \dots, T-1\}$, and $T > h$ ensures sufficient history. The objective optimizes $v_t^\phi$ by regressing a target velocity vector field $\dot{x}_t$ constructed on paired consecutive points, explicitly:
where the target velocity field is constructed via linear interpolation in normalized time between current and next points: $x_t = (1 - t) x_k + t x_{k+1}$, thus $\dot{x}_t = \frac{d}{dt}x_t = x_{k+1} - x_{k}$ for $ t \in [0,1]$. 
Incorporating the physics model $f_p$ into the model and objective yields the following GB-DM loss:
\begin{align}
\label{eq:GB-Dyn}
& \mathcal{L}_{\text{GB-DM}}(\phi) = \notag \\
&  \mathbb{E}_{\substack{\vect{x} \sim \pi(\vect{x}) \\ x_{k-h:k+1} \sim \pi(\cdot | \vect{x}) \\ t \sim \mathcal{U}(0,1) }}
    \left\| (v_t^\phi (x_t \mid \vect x_{k-h:k}) \circ f_p (x_t, \cdot ))
    - \dot{x}_t \right\|^2,
\end{align}
where $x_t$ describes the interpolant built between data points, and $\dot{x}_t$ is the associated time derivative \footnote{We use the generic term of interpolant, which can be linear and non-linear. An example of the latter one is later introduced for solving II order ODE/PDEs}.
The physics model $f_p$ is combined with a learnable vector field $v_t^\phi$, representing the missing dynamics $f_\phi$, effectively learning to complete the physics dynamics by aligning the combined field with the target velocity constructed by the data.
\paragraph{Structured Variational Inference.}
A key challenge in incorporating the physics model $f_p (x_t, \theta)$ is that the physics parameters $\theta \in \Theta \subseteq  \mathbb{R}^p $ are not provided in the data. Furthermore, purely deterministic physics models often fail to capture stochasticity or multimodal-behaviour that exists in many dynamic systems. This multi-modality manifests in a velocity vector field as multiple plausible flow directions at a single point in state-space. This is a common phenomenon in grey box modelling where unknown physics can lead to divergent trajectories from identical initial conditions.

To address these, we introduce a structured latent variable approach. We introduce a latent variable $\theta$ to represent the physics parameters and a second latent variable $z \in \mathcal{Z} \subseteq \mathbb{R}^h$ to capture the missing stochasticity and multi-modal behaviour in the velocity vector field. 

We model the distribution of the vector field variationally by introducing a variational posterior conditioned on the trajectory segment $q_{\psi}(z, \theta \mid \vect{x}_{k-h:k})$. A probabilistic schema of this latent variable structure is provided in Figure \ref{fig:PGM}.
\begin{align}
    \begin{split}
\log & \; p_t^\phi(\dot{x}_t \mid x_k) 
    = \log \int p_t^\phi(\dot{x}_t, \theta, z \mid x_k)\, d\theta\, dz \\
    &= \log \int q_{\psi}(z, \theta \mid \vect{x}_{k-h:k}) \frac{p_t^\phi(\dot{x}_t, \theta, z \mid x_k)}{q_{\psi}(z, \theta \mid \vect{x}_{k-h:k})}\, d\theta\, dz
    \end{split}
    \label{eq_marginalizing}
\end{align}
We assume a structured factorization for the variational posterior, where the latent variable that corresponds to the physical parameters depends on the stochasticity latent variable: $q_{\psi}(z, \theta \mid \vect{x}_{k-h:k})=q_{\psi}(\theta \mid \vect{x}_{k-h:k}, z) q_{\psi}(z \mid \vect{x}_{k-h:k})$. Applying Jensen's Inequality to \eqref{eq_marginalizing}, we derive a variational dynamics matching objective similarly to \cite{guo2025variationalrectifiedflowmatching}:
\begin{align}
\label{eq:ELBO}
    \log p_t^\phi(\dot{x}_t \mid x_k) & \ge \mathbb{E}_{q_{\psi}(z, \theta \mid \vect{x}_{k-h:k})}[\log p_t^\phi(\dot{x}_t \mid \theta, z, x_k)] \nonumber \\
    & \hspace{-1cm}\nonumber - \mathbb{E}_{q_{\psi}(z |\vect{x}_{k-h:k})} \text{KL}\left[q_{\psi}(\theta \mid \vect{x}_{k-h:k}, z) \Vert p(\theta)\right]  \\
    & \hspace{-1cm } -\text{KL}\left[q_{\psi}(z \mid \vect{x}_{k-h:k}) \Vert p(z)\right],
\end{align}
where $p(\theta)$ is the physics informed prior and $p(z)=\mathcal{N}(\vect 0, \mathbf{I})$.
Assuming a Gaussian likelihood for the velocity matching  (corresponding to the squared error), \cite{guo2025variationalrectifiedflowmatching}, our objective becomes:
\begin{align}
\label{eq:GB-V-DynM}
    &\mathcal{L}_{\text{VGB-DM}}^{\text{VI}}(\phi, \psi)= \mathbb{E}_{\substack{\pi(\vect{x}),\pi(\cdot | \vect{x})}} \Bigl[ \Bigr. \notag \\
    & \quad - \mathbb{E}_{\substack{q_\psi(z, \theta \mid \vect{x}_{k-h:k}) \\ t \sim \mathcal{U}(0,1)}} \Big\| (v_t^\phi(x_t \mid  \theta, z) \circ f_p(x_t, \theta)) 
    - \dot{x}_t \Big\|^2 \notag \\  
    & \quad \Bigl. - \text{KL}\left[q_{\psi}(\theta, z \mid \vect{x}_{k-h:k}) \Vert p(\theta,z)\right] \Bigr]
\end{align}
In Appendix \ref{app:proofs}, we provide the derivation of the ELBO and a proof that the learned distribution preserves the marginal data distribution. Practically, we model the approximate posterior by an encoder such that $q_\psi (z | \vect{x}_{k-h:k})=\mathcal{N}\bigl(z; \mu_{\psi}(\vect{x}_{k-h:k}), \sigma_\psi(\vect{x}_{k-h:k})\bigr)$ and $q_\psi (\theta | \vect{x}_{k-h:k}, z)=\mathcal{N}\bigl(\theta; \mu_{\psi}^{'}(\vect{x}_{k-h:k}, z), \sigma_\psi^{'}(\vect{x}_{k-h:k},z)\bigr)$, enabling analytic computation of the KL-divergence in \eqref{eq:GB-V-DynM}. We illustrate a schematic view of training and inference of our model in Figure \ref{fig:main-diagram}.

\paragraph{Second-Order Dynamics.}
Physics systems can also include second-order dynamics, described by equations in the form of $\frac{\partial^2 x_t}{\partial t^2}= f_p(x_t, \theta)$, such as those governing pendulums, wave equations, and similar systems. To integrate these physics systems, we extend our objective 
\eqref{eq:GB-V-DynM} to this setting, by introducing a higher order, a target acceleration vector field. 
Specifically, we regress both velocity and acceleration vector field by a non-linear interpolation $\mathcal{I}(x_t; x_{k-1}, x_k, x_{k+1})$ using three consecutive trajectory points, retrieving its velocity and acceleration vector fields: $\frac{\partial\mathcal{I}}{dt}=\dot{x}_t$ and $\frac{\partial^2\mathcal{I}}{\partial t^2}=\ddot{x}_t$. The learned dynamics are 
then modelled as the paired system of velocity and acceleration fields
$$
u_t^\phi(x_t,\dot{x}_t \mid \theta,z)
:= 
\begin{bmatrix}
v_t^\phi(x_t \mid \theta,z) \\
a_t^\phi(x_t,\dot{x}_t \mid \theta,z) \circ f_p(x_t,\theta)
\end{bmatrix}.
$$
We modify the vector field matching term in \eqref{eq:GB-V-DynM} to obtain the second-order objective:
\begin{align}
\label{eq:second-order-dm}
&\mathbb{E}_{\substack{(z,\theta)\sim q_\psi(z,\theta\mid\vect{x}_{k-h:k}) \\ t \sim \mathcal{U}(0,1) }} \left[
\|v_t^\phi(x_t \mid \theta,z) - \dot{x}_t\|^2 \right. \nonumber \\ 
& \quad \left.  + \; \alpha \; \|a_t^\phi(x_t,\dot{x}_t \mid \theta,z)\circ f_p(x_t,\theta) - \ddot{x}_t\|^2
\right],
\end{align}
where $\alpha$ is a hyper-parameter that balances the optimization of the acceleration term to prevent excessive smoothing of the trajectory. In practice, we implement the non-linear interpolation using Lagrange interpolation.\footnote{Other interpolation methods, such as Fourier interpolation or spline-based approaches, could be explored, but they are beyond the scope of this work.} Additionally, the velocity and acceleration vector fields $u^\phi_t$ share a backbone network, with the velocity and acceleration modelled as two distinct heads of the network. This architecture leverages shared parameters to ensure consistency between the learned velocity and acceleration fields. We implement this method when solving the Pendulum system as reported in our experiment, setting $\alpha=0.5$.

\section{Related works} \label{sec:related_works}
We position our Variational Grey-Box Dynamics Matching (VGB-DM) framework within the landscape of learning dynamical systems,  highlighting the distinctive features of our algorithm compared to existing data-driven and grey-box approaches.

Fully data-driven approaches such as Neural ODEs \cite{chen2018neuralode} and Flow Matching-based \cite{lipman2023, albergo2023, liu2023flow} including adaptation for forecasting time-series as in \cite{zhang2024tjfm} demonstrate modelling complex distributions using data, but operate as black boxes. However, these solutions can face limitations when training data are scarce, struggles when extrapolating beyond observed regimes, and offer limited interpretability.
By contrast, VGB-DM incorporates incomplete physics models, enabling efficient learning even with limited data while maintaining interpretability through explicit physical parameters that can be inferred. Moreover, they enable further mechanistic inspection of the physics-informed and the learnable components, as studied by \cite{naoya_2023,singh2025hybrid} analysing physics regularizers.

Recent grey-box modelling approaches build on Neural ODEs by combining physics-based models with data-driven neural networks, achieving strong results in parameter inference and dynamics forecasting \cite{yin_2021, naoya_2021, Mehta2020NeuralDS, qian2021pharma_ode, wehenkel2023robust, verma2024climode}. However, these simulation-based methods require numerical ODE solvers during training along with physics model dynamics, presenting scaling difficulties for high-dimensional settings due to gradient computation through ODE solvers and parameter updates, particularly for long time horizons or complex dynamics leading to training difficulties and numerical stability \cite{chen2018neuralode,naoya_2021,wehenkel2023robust,yin_2021}.
VGB-DM bypasses this paradigm by adopting a simulation-free method, avoiding these computational bottlenecks and stability issues, enabling more scalable and robust grey-box modelling for high-dimensional systems.

Notably, the work on Variational Rectified Flow Matching \cite{guo2025variationalrectifiedflowmatching} identifies a key limitation of deterministic flow matching methods: they struggle when modelling multi-modal velocity fields because the standard flow matching objective leads to averaging over multiple target directions of the velocity vector field. To overcome this issue, they introduce a variational posterior $q_\psi(z | x_t, t)$ over latent variables, providing the necessary flexibility to model multimodal velocity distributions. However, VGB-DM introduces a structured variational posterior tailored for grey-box modelling and explicitly separates latent variables into two components: one capturing stochasticity and multi-modality in the dynamics, and another inferring interpretable physical parameters. This structured approach enhances both the expressiveness of the learned dynamics and the interpretability of the inferred parameters.

Finally, to integrate second-order dynamics systems, we extend our simulation-free method using non-linear interpolants (e.g. Lagrange Interpolation). This enables direct regression of both velocity and acceleration fields, addressing a broader class of physical systems than existing flow matching approaches.

\section{Experiments} \label{sect:exps}
We conduct experiments on both synthetic datasets derived from differential equations (PDE/ODEs) (Section~\ref{sec:ode_datasets}) and a real-world weather modelling task (Section~\ref{sec:weather_dataset}). All experiments are repeated with multiple random seeds to ensure consistency and evaluate the stability of the compared methods\footnote{All our models were trained on a single RTX 3060 (12 GB), while ClimODE was trained on an RTX A5000 and required at least 25 GB.}.

\begin{table*}[t]
\centering
\small
\begin{tabular}{lcccc}
\toprule
Model & \begin{tabular}{c}Pendulum \\ ($\times 10^{-2}$)\end{tabular} & 
       \begin{tabular}{c} RLC \\ ($\times 10^{-1}$)\end{tabular} & 
       \begin{tabular}{c} Reaction Diffusion \\ ($\times 10^{-3}$)\end{tabular} &
       \begin{tabular}{c} Lorenz \\ ($\times 10^{-1}$)\end{tabular} \\
\midrule
PhysVAE~\cite{naoya_2021}   & $0.417 \pm 0.001$          & $0.29 \pm 0.01$          & $7.4 \pm 1.3$  & $13.12 \pm 1.27$\\
VBB-DM~\cite{guo2025variationalrectifiedflowmatching}     & $1.058 \pm 0.004$          & $0.38 \pm 0.03$          & $8.0 \pm 1.1$  & $9.91 \pm 1.22$\\
BB-NODE~\cite{chen2018neuralode}   & $0.452 \pm 0.001$          & $0.39 \pm 0.04$          & $7.5 \pm 0.5$   & $22.15 \pm 7.84$\\
TFM~\cite{zhang2024tjfm}       & $0.348 \pm 0.001$          & $0.26 \pm 0.01$          & $8.5 \pm 2.8$ & $9.85 \pm 1.32$ \\ \midrule
Ours      & $\mathbf{0.283} \pm 0.001$ & $\textbf{0.23} \pm 0.01$ & $\mathbf{7.1} \pm 0.9$  & $\mathbf{4.68} \pm 0.16$ \\ 
\bottomrule
\end{tabular}
\caption{\textbf{Forecasting performance comparison (MSE).} Results on four dynamical systems benchmarks. Our method achieves the lowest error across all tasks.}
\label{tab:forecast_mse}
\end{table*}

\subsection{Differential-Equation Datasets} \label{sec:ode_datasets}

\textbf{Experimental setup.}
We evaluate our method on four canonical dynamical systems governed by differential equations: an RLC circuit, a reaction–diffusion (RD) system, a damped pendulum, and a chaotic Lorenz attractor system. In all experiments, grey-box models have access only to incomplete physics equations, while the dataset consists of state trajectories generated from the complete governing equations, following the experimental protocol of ~\cite{naoya_2021, wehenkel2023robust}. Details of dataset construction and model architectures (encoder and vector field architectures) are provided in Appendix ~\ref{sec:data_generation}. 
For each experiment, we follow the same network architecture design proposed in previously established benchmarks of grey-box models, for PhysVAE, this includes the additional regularization and hyperparameters suggested in ~\cite{naoya_2021, wehenkel2023robust, yin_2021}.

\textbf{Baselines.}
For comparison, we benchmark against four baselines: PhysVAE ~\cite{naoya_2021}, a state-of-the-art grey-box method, and two widely adopted black-box approaches where physics is not included in the model nor in the objective: Neural ODE for time series ~\cite{chen2018neuralode} (BB-NODE), the Variational Rectified Flow Matching ~\cite{guo2025variationalrectifiedflowmatching} adapted for time series (VBB-DM) and Trajectory Flow Matching (TFM) \cite{zhang2024tjfm}.
All models are compared with equivalent network architecture complexity to ensure fair evaluation among them. 

\textbf{Forecasting performance.}
Table~\ref{tab:forecast_mse} presents mean squared error (MSE) results for time-series forecasting, where each model was trained on the largest training set size for each task. To assess extrapolation capabilities and generalization beyond the training regime, forecasts were tested on horizons three or four times longer than those encountered during training.
Across all four benchmarks, our method consistently achieves lower errors than all baselines, demonstrating superior predictive accuracy and highlighting the benefit of the incomplete physics model as an inductive bias for learning the dynamics. In addition to the trajectory forecasts, Figure \ref{fig:params-identification} of Appendix \ref{app:addition-results} also reports the physics parameter estimation errors of the grey-box models, as part of our interpretability analysis.

\begin{figure*}[t]
\centering
\includegraphics[width=0.32\textwidth]{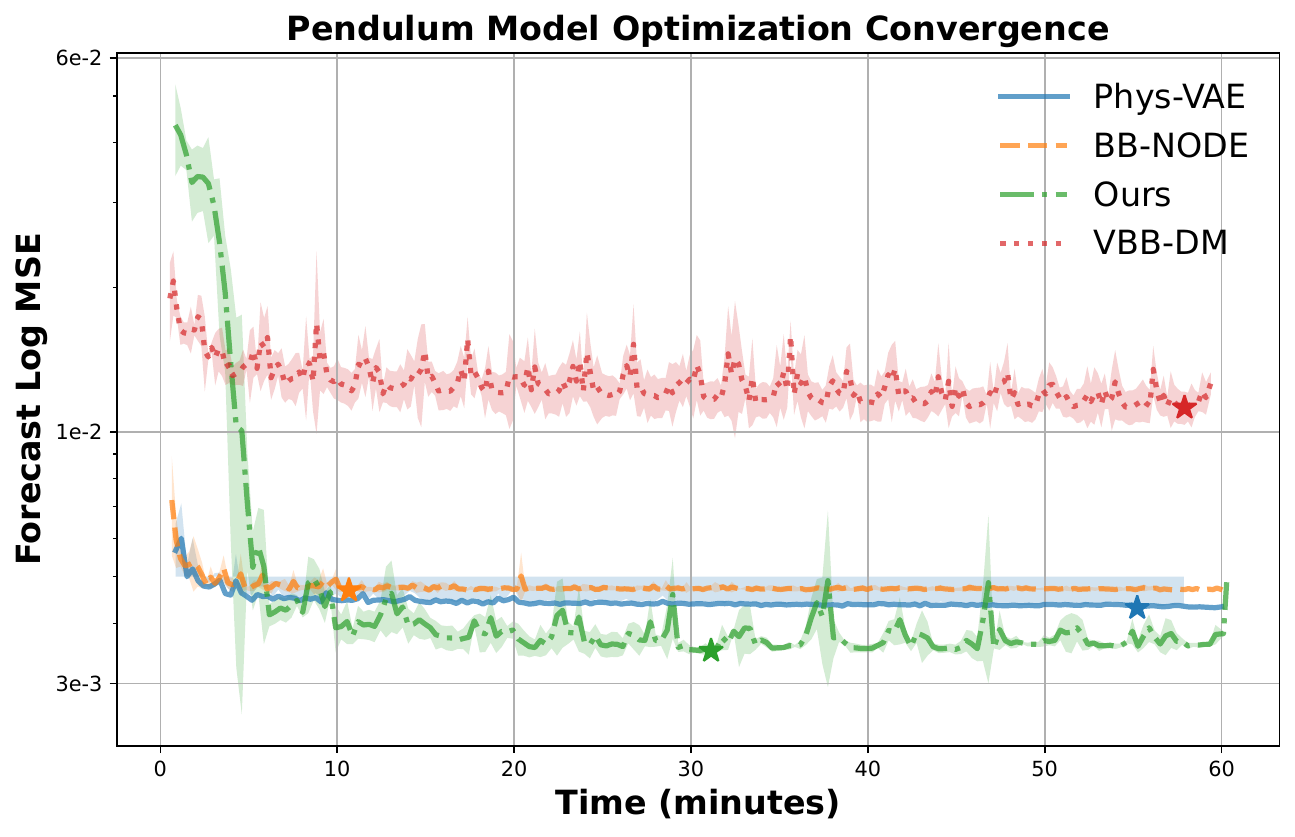}
\includegraphics[width=0.32\textwidth]{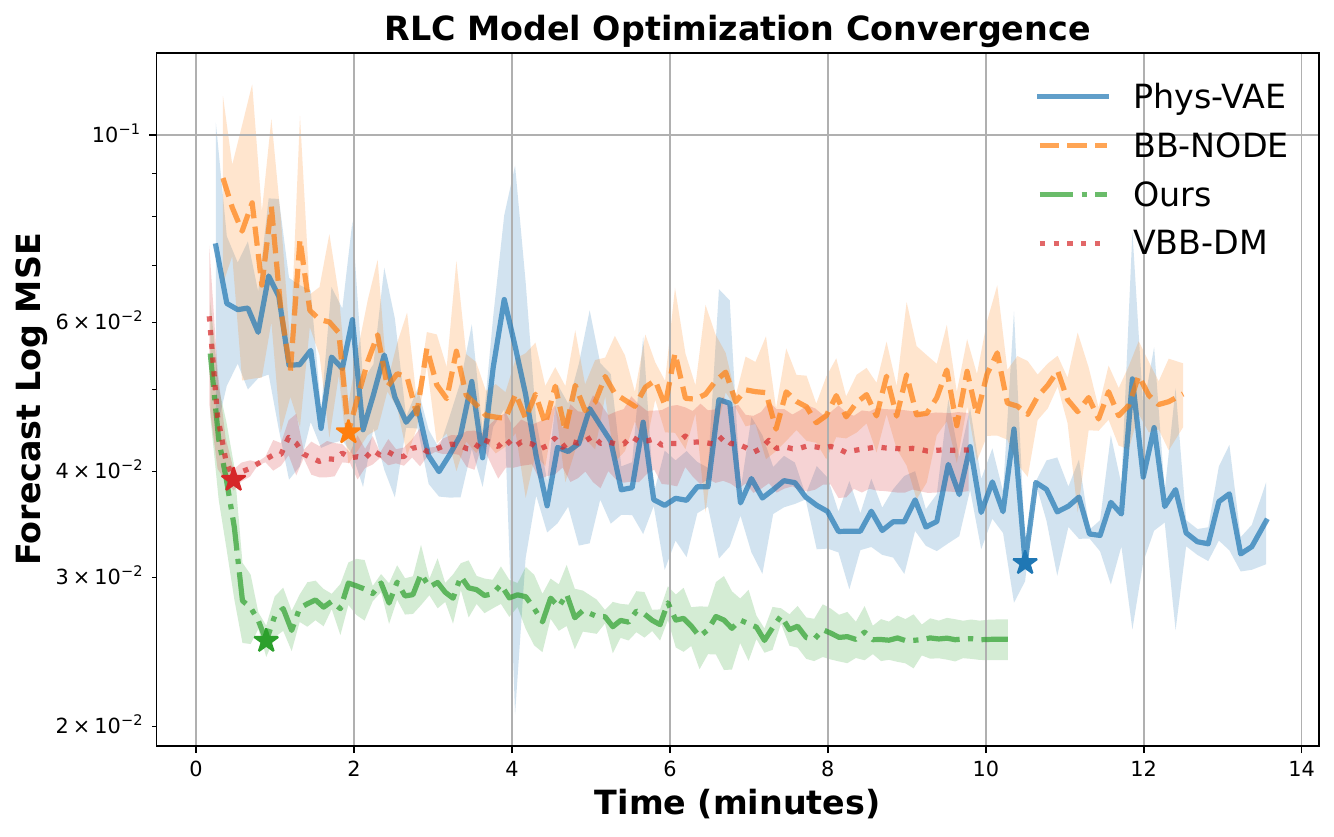}
\includegraphics[width=0.32\textwidth]{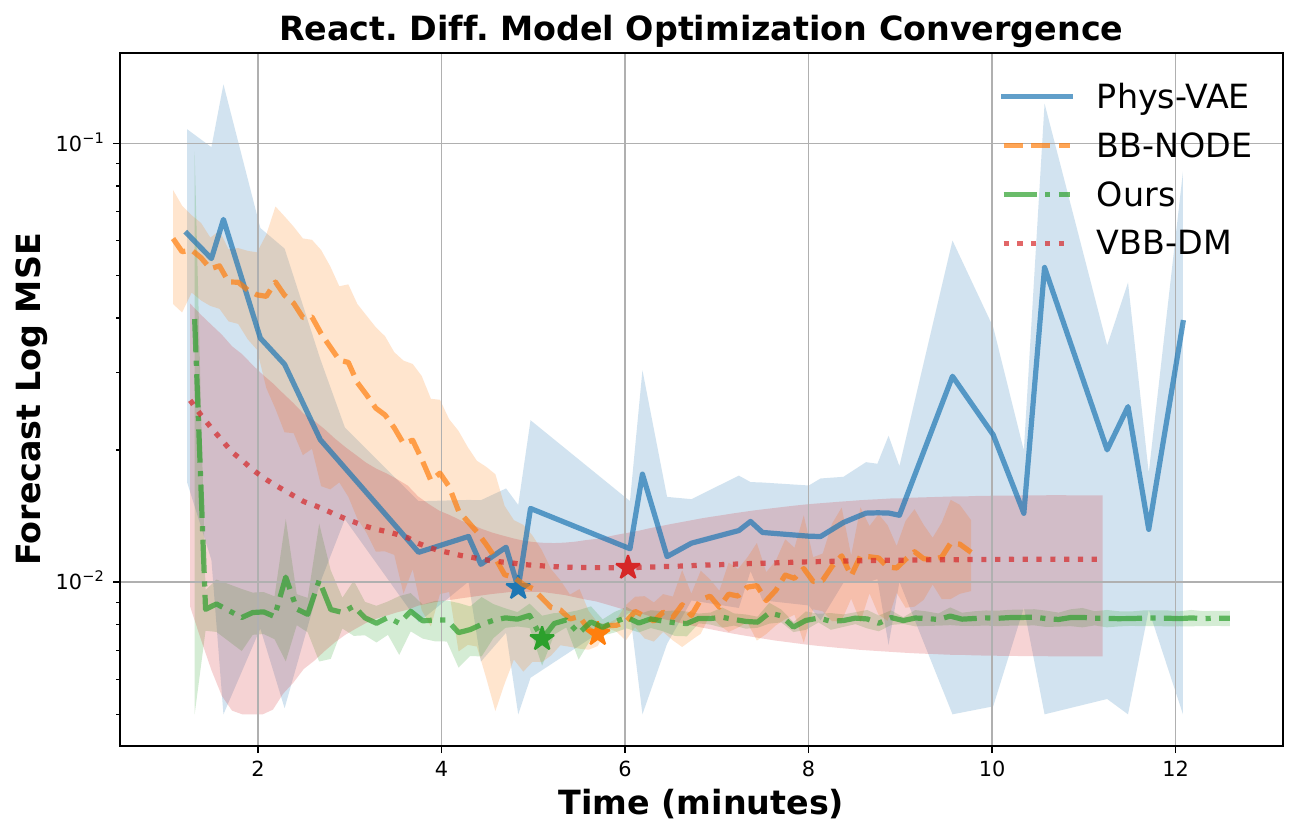}
\caption{\textbf{Optimization convergence analysis.} Forecast $\log$MSE vs.\ training time (minutes) for the Pendulum (left), RLC (center), and Reaction-Diffusion (right) tasks using the largest training dataset size. Our method demonstrates significantly faster convergence and achieves lower final error compared to simulation-based methods (PhysVAE, BB-NODE), while also exhibiting better stability than the black-box dynamics matching baseline (VBB-DM). This highlights the benefit of avoiding backpropagation through the ODE solver.}
\label{fig:opti-convergence}
\end{figure*}

\textbf{Convergence behaviour.}
In Figure~\ref{fig:opti-convergence}, we show the training convergence of each method on the test set across three training seeds. Our approach converges rapidly and attains lower final error than competing methods. By leveraging a simulation-free training objective, it avoids backpropagation through the ODE solver, which is a known source of gradient instability in simulation-based approaches such as PhysVAE and BB-NODE. This results in smoother and more consistent optimization trajectories. Furthermore, since each training step does not require running a numerical integrator, our method is significantly cheaper per iteration, translating into faster wall-clock convergence.

\begin{figure*}[t]
\centering
\includegraphics[width=0.98\textwidth]{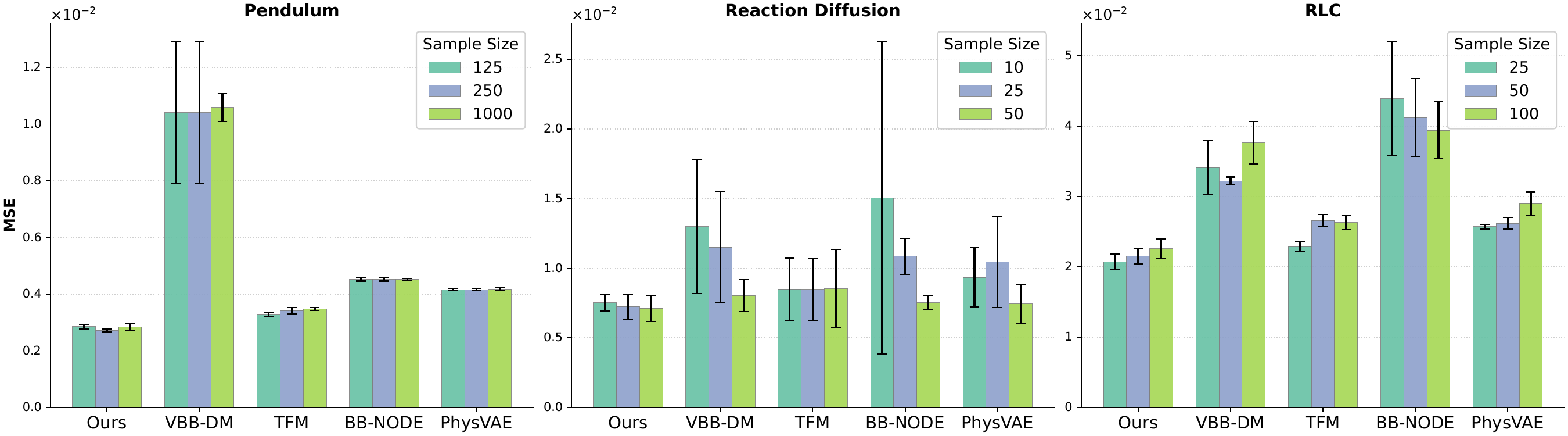}
\caption{\textbf{Sample efficiency analysis.} Forecast MSE across varying training sample sizes for Pendulum, RLC, and Reaction-Diffusion tasks. Grey-box methods (Ours and PhysVAE) consistently outperform their black-box counterparts (VBB-DM and BB-NODE) across all regimes, with our method achieving the lowest error and variance, particularly under limited data.}
\label{fig:sample_size}
\end{figure*}

\begin{figure*}[t]
\centering
\includegraphics[width=1.0\textwidth, height=6cm]{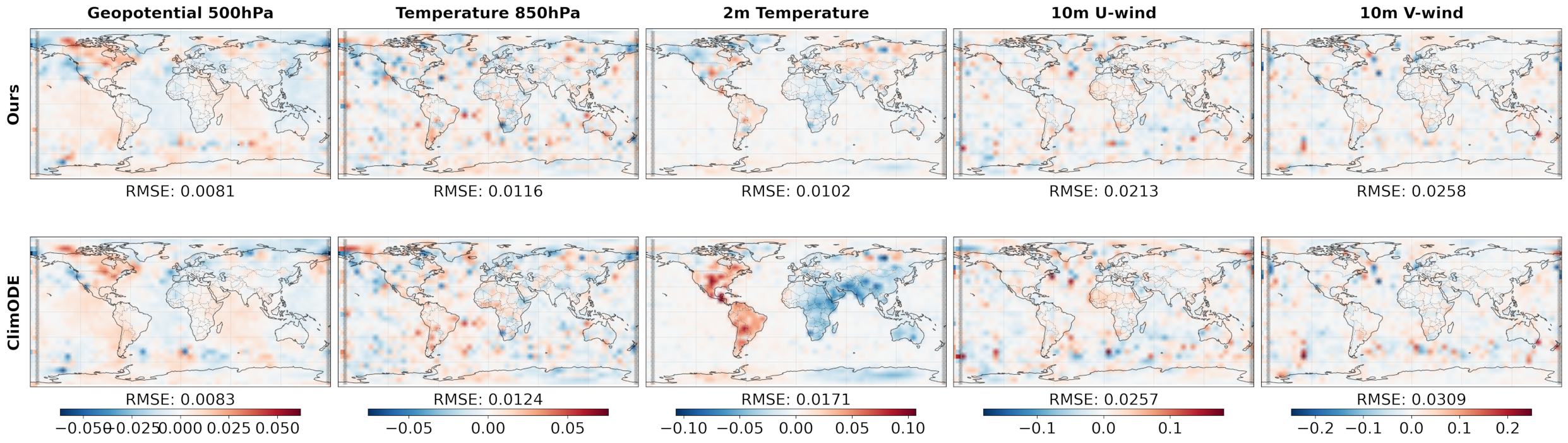}
\caption{\textbf{Residual maps for weather forecasting} Visualization of absolute error maps between ground truth and predicted fields for five meteorological variables (Ours vs ClimODE)}
\label{fig:residual_mse}
\end{figure*}

\begin{figure*}[t] 
\centering
\includegraphics[width=0.999\textwidth,]{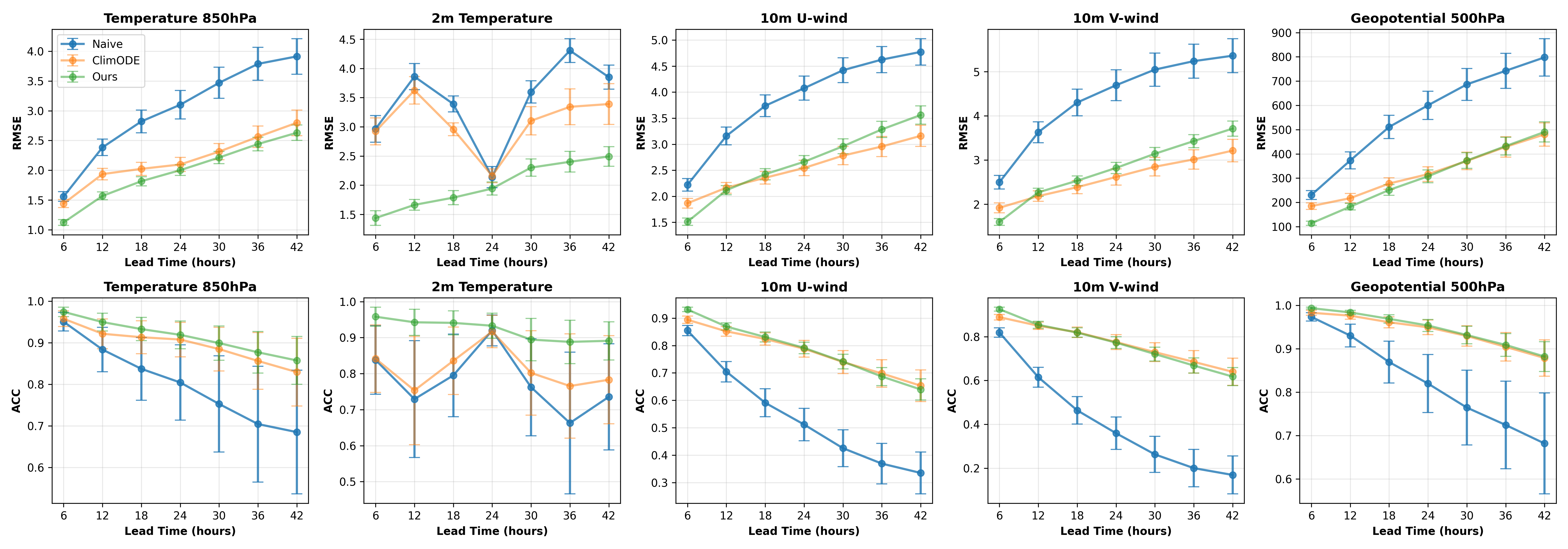}
\caption{\textbf{Forecasting performance over $42$ hours.} Latitude-weighted RMSE (top, lower is better) and ACC (bottom, higher is better) for ERA5 dataset. Our method performs better or on par than ClimODE.}
\label{fig:clime_mse_global}
\end{figure*}

\textbf{Sample efficiency.}
Grey-box approaches are often credited with providing robust performance under limited data~\cite{wehenkel2023robust, yin_2021,naoya_2021,naoya_2023}.
We evaluate each method across three training set sizes, ranging from 10 to 1000 samples, in order to assess robustness to data scarcity.
In Figure~\ref{fig:sample_size}, both grey-box models (ours and PhysVAE) consistently outperform their black-box counterparts: VGB-DM (ours) vs. VBB-DM and Phys-VAE vs. BB-NODE. Our method achieved lower error with smaller variance across all regimes, including when compared to TFM. Moreover, it further improves upon PhysVAE, exhibiting more stable predictions and reduced sensitivity to sample size. In contrast, VBB-DM and BB-NODE display substantially higher variance, especially in low-data regimes, suggesting that the physical prior in the grey-box models plays a critical regularization role when training data is scarce. These results confirm that embedding structural knowledge into the model is particularly advantageous when data collection is costly or limited.

\subsection{Weather modelling} \label{sec:weather_dataset}
Physics-based formulations can provide valuable inductive biases for real-world datasets. In particular, large-scale weather dynamics can be modelled using the Advection equation, as demonstrated by \cite{verma2024climode}. To assess the effectiveness of our method, we conduct experiments on the ERA5 reanalysis dataset for medium-range weather forecasting. The dataset includes five key meteorological variables:  
\textit{(i)} geopotential,  
\textit{(ii)} ground temperature,  
\textit{(iii)} atmospheric temperature,  
\textit{(iv)} and \textit{(v)} the two ground-level wind components (u10 and v10).  

\textbf{Experimental Setup.}  
We adopt the preprocessing pipeline and train/test splits introduced in ClimODE~\cite{verma2024climode}, and evaluate performance at two temporal resolutions: hourly and monthly. To ensure a fair comparison, we train our model and ClimODE under identical conditions, using the same maximum number of optimization steps (2000) and a fixed time budget of six hours.  Evaluation is based on two standard metrics: latitude-weighted Root Mean Squared Error (RMSE) and the Anomaly Correlation Coefficient (ACC). Our goal is to provide a simulation-free alternative to physics-informed neural ODEs; hence, we restrict our comparison to ClimODE. We also include a naive persistence baseline (last observation) as a reference point to contextualize the results. Additional implementation and training details are provided in Appendix ~\ref{app:details_weather_exp}.

\textbf{Residual maps for hourly resolution.}  
Figure~\ref{fig:residual_mse} visualizes the absolute error maps and per-feature RMSE for a representative test sample. Across all five variables, our method yields lighter residual structures compared to ClimODE, particularly over oceanic regions. This indicates systematically reduced errors, which is further corroborated by consistently lower RMSE values in quantitative evaluations.

\begin{figure*}[t] 
\centering
\includegraphics[width=0.98\textwidth]{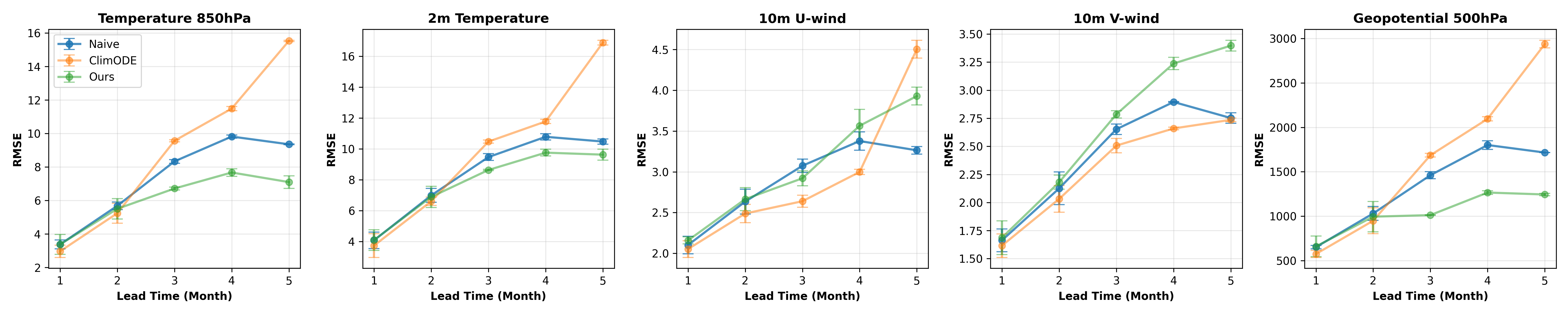}
\caption{\textbf{Long-term forecasting performance over $5$ months.} Latitude-weighted RMSE for ERA5 dataset over an extended period. Our method outperforms ClimODE by mitigating compounding errors over long horizons, with the performance gap widening as the forecast horizon extends.}

\label{fig:clime_mse_month}
\end{figure*}

\textbf{Results for hourly resolution.} 
Figure~\ref{fig:clime_mse_global} presents RMSE and ACC values over a 42-hour forecast horizon. Our approach performs on par with or better than ClimODE, achieving higher ACC and lower RMSE across the time steps.

\textbf{Results for monthly resolution.} 
Figure~\ref{fig:clime_mse_month} extends the evaluation to a 5-month forecast horizon. The advantage becomes more pronounced as the forecast horizon extends, indicating that our method retains stronger temporal stability compared to Neural ODE–based formulations. This aligns with our earlier experiments on synthetic ODE datasets, where our method also demonstrated greater robustness over extended prediction horizons, underscoring its ability to mitigate compounding errors — a challenge that traditional Neural ODE models struggle to overcome.

\section{Conclusion}
We introduced Variational Grey-Box Dynamics Matching (VGB-DM), a novel framework that combines the interpretability of physics-based models with the efficiency of simulation-free generative learning. By adopting a simulation-free training paradigm, VGB-DM bypasses computational and memory bottlenecks and stability issues inherent in traditional solver-based grey-box methods. Our structured variational approach effectively disentangles physical parameters from unknown stochasticity, enabling robust dynamics learning and interpretable parameter inference. Empirical evaluations on various ODE/PDE systems and a real-world weather forecasting task demonstrate that VGB-DM achieves state-of-the-art forecasting performance and faster convergence than existing grey-box and black-box baselines, crucially, while preserving the interpretability of the underlying physics model.
\paragraph{Limitations and Future Work.} While VGB-DM offers significant advantages, it relies on several assumptions. First, the framework assumes that the underlying dynamics and the physics model are differentiable and relatively smooth. The performance on highly stiff or discontinuous physical systems remains an area for future investigation. However, the simulation-free nature of VGB-DM avoids gradient instability issues often encountered when backpropagating through solvers applied to stiff systems. Beyond these considerations, an important direction for future work is transfer across related physical systems. In similar domain (e.g., from a single to a double pendulum), the encoder can be reused as a pre-trained feature extractor, with only the output layers adapted if physical parameters are different. In this setting, a pre-trained VGB-DM model provides a strong warm start and can be fine-tuned to capture additional latent components arising from richer dynamics, enabling efficient knowledge transfer across families of systems.

\subsubsection*{Acknowledgements}
We acknowledge the financial support of the Swiss National Science Foundation (SNF). G. Mercatali has been supported by SNF project IZLJZ2\_214000, Interpretable Condition Monitoring, F. Lavda has been supported by SNF 200021\_207428, Learning generative models for molecules, G. Singh SNF project CRSII5\_209434, Migrate, A Multidisciplinary and InteGRated Approach for geoThermal Exploration. The computations were performed at the University of Geneva on "Baobab" and "Yggdrasil" HPC clusters.

\bibliography{bibliography}

\section*{Checklist}



\begin{enumerate}

  \item For all models and algorithms presented, check if you include:
  \begin{enumerate}
    \item A clear description of the mathematical setting, assumptions, algorithm, and/or model. [Yes] See Section ~\ref{sec:problem_settings} and ~\ref{sec:method}.
    \item An analysis of the properties and complexity (time, space, sample size) of any algorithm. [Yes] See Section~\ref{sec:method}
    \item (Optional) Anonymized source code, with specification of all dependencies, including external libraries. [Yes] Will release upon acceptance.
  \end{enumerate}

  \item For any theoretical claim, check if you include:
  \begin{enumerate}
    \item Statements of the full set of assumptions of all theoretical results. [Yes] See Sections~\ref{sec:background} and \ref{sec:method}
    \item Complete proofs of all theoretical results. [Yes] Appendix ~\ref{app:proofs}.
    \item Clear explanations of any assumptions. [Yes] See Section~\ref{sec:problem_settings}
  \end{enumerate}

  \item For all figures and tables that present empirical results, check if you include:
  \begin{enumerate}
    \item The code, data, and instructions needed to reproduce the main experimental results (either in the supplemental material or as a URL). [Yes] We report in Appendix \ref{app:ode_dataset_details} for ode-datasets and in Appendix  ~\ref{app:details_weather_exp}.
    \item All the training details (e.g., data splits, hyperparameters, how they were chosen). [Yes] Appendix~\ref{sec:ode_datasets} for ode-datasets and in Appendix ~\ref{app:details_weather_exp}.
    \item A clear definition of the specific measure or statistics and error bars (e.g., with respect to the random seed after running experiments multiple times). [Yes] In Section~\ref{sect:exps}
    \item A description of the computing infrastructure used. (e.g., type of GPUs, internal cluster, or cloud provider). [Yes] In Section~\ref{sect:exps}.
  \end{enumerate}

  \item If you are using existing assets (e.g., code, data, models) or curating/releasing new assets, check if you include:
  \begin{enumerate}
    \item Citations of the creator If your work uses existing assets. [Yes]
    \item The license information of the assets, if applicable. [Yes]
    \item New assets either in the supplemental material or as a URL, if applicable. [Yes]
    \item Information about consent from data providers/curators. [Not Applicable] Not used.
    \item Discussion of sensible content if applicable, e.g., personally identifiable information or offensive content. [Not Applicable] Not used.
  \end{enumerate}

  \item If you used crowdsourcing or conducted research with human subjects, check if you include:
  \begin{enumerate}
    \item The full text of instructions given to participants and screenshots. [Not Applicable] Not used.
    \item Descriptions of potential participant risks, with links to Institutional Review Board (IRB) approvals if applicable. [Not Applicable] Not used.
    \item The estimated hourly wage paid to participants and the total amount spent on participant compensation. [Not Applicable] Not used.
  \end{enumerate}

\end{enumerate}

\clearpage

\appendix
\thispagestyle{empty}
\onecolumn
\aistatstitle{Supplementary Materials}



\begin{figure*}[h]
\centering
\includegraphics[width=0.99\textwidth]{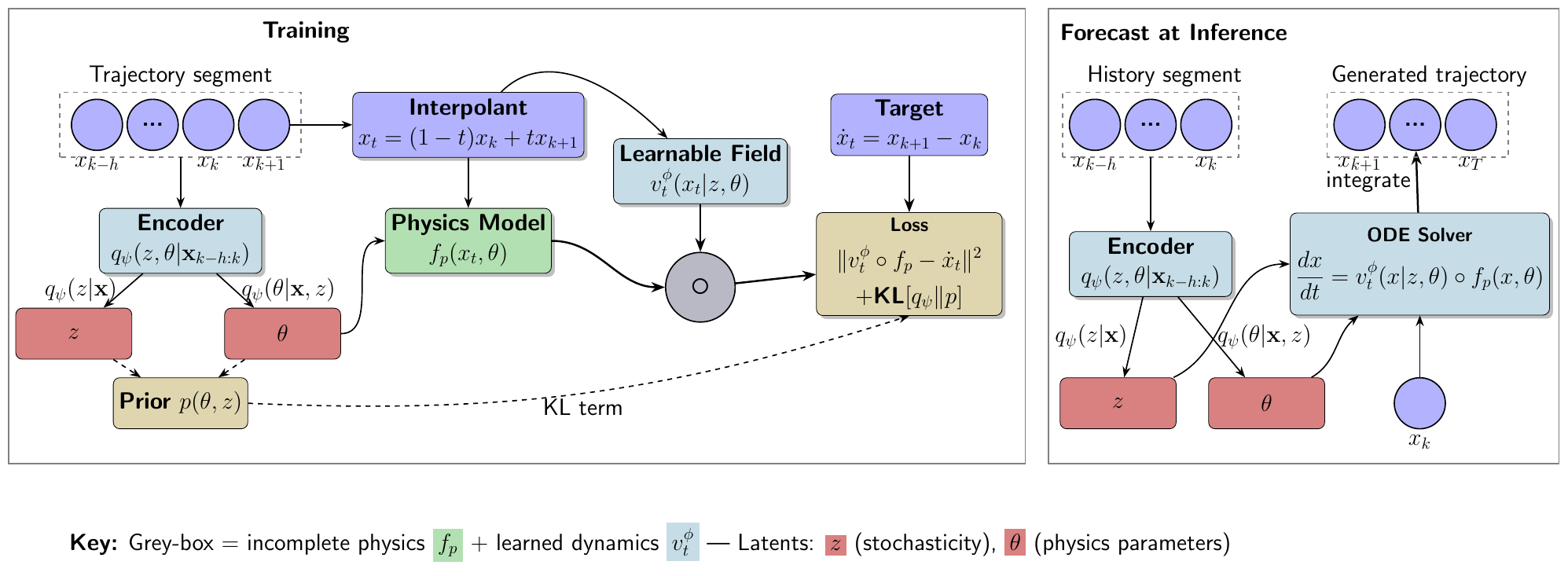}
\caption{Overview of the Variational Grey-Box Dynamics Matching (VGB-DM) framework.}
\label{fig:diagram}
\end{figure*}

\paragraph{Method Overview.}
Figure~\ref{fig:diagram} illustrates the proposed \textbf{Variational Grey-Box Dynamics Matching (VGB-DM)} framework. 
During training, the model is \emph{simulation-free}: it does not require integrating the dynamics to compute gradients. 
Instead, the physics-based component $f_p$ is directly combined with the learnable vector field $v_t^{\phi}$. The training objective is optimized without numerical simulations, avoiding memory and computation bottlenecks of backpropagation through the solver and optimization instability.
The model includes an encoder that maps a segment of the observed trajectory into two structured latent variables: 
(i) the stochastic latent variable $z$, and 
(ii) the physics-related variable $\theta$. 
These latents condition both the physics model and the learnable field, allowing the system to infer underlying physical parameters and represent multimodal velocity vector fields.

At inference (forecasting) time, the model predicts future trajectories by integrating the combined dynamics of the physics model and the learnt component. Multiple forecast realisations can be generated by sampling different latents from the learnt posterior.

\section{Differential-Equation Benchmarks} \label{app:ode_dataset_details}
The synthetic dataset is based on Differential Equation, where initial conditions and parameters are stochastic, following the work of \cite{naoya_2021, wehenkel2023robust}.
In the next subsections we report each experiment describing data generation and model architecture and hyperparameters used.
For each experiment, we followed the same networks architecture design proposed by ~\cite{naoya_2021, wehenkel2023robust}.
We compare our method with its black-box versions using the simulation-based methods (BB-NODE) and the simulation-free methods (VBB-DM), where in the latter physics is not included in the model. In addition to our grey-box model we also compared its performance with the Phys-VAE \cite{naoya_2021} used also by \cite{wehenkel2023robust}. We follow for each experiment all the hyperparameters used by the models and regularisers as listed in their work.

\subsection{Experiments and implementations} \label{sec:data_generation}
\paragraph{RLC.} The RLC series circuit, consisting of a resistor (R), capacitor (C), and inductor (L), models a broad class of transfer functions. The system state at time $t$ is defined as $x_t = [I_t, U_t]^\top$, where $U_t$ denotes the capacitor voltage and $I_t$ the circuit current. The state evolution follows: 
\begin{align} \frac{\partial}{\partial t} \begin{bmatrix} I_t \ U_t  \end{bmatrix} = \underbrace{\begin{bmatrix} \frac{I_t}{C} \ \frac{1}{L}(V_t - U_t) \end{bmatrix}}_{f_p} + \underbrace{\begin{bmatrix} 0 \ -\frac{R}{L}I_t \end{bmatrix}}_{\text{missing physics}}. 
\end{align}
The dataset is generated by sampling $L$ and $C$ uniformly from $[1, 3] \times [0.5, 1.5]$, and $R \sim \mathcal{U}(1, 3)$. The input voltage $V_t$ is a fixed, known signal. Initial conditions are given by $U_0 \sim \mathcal{N}(0, 1)$ and $I_0 = 0$. The incomplete physics model $f_p$ in the grey-box setup includes only the first term of the full equation.

Each trajectory is simulated from $t_0 = 0$ to $t_{\text{max}} = 20$ seconds with a time resolution of $\Delta t = 0.1$ s, yielding 200 time points per trajectory. The training, validation, and test sets contain 1000, 100, and 100 samples, respectively. During training, trajectories are divided into segments (history windows) of size $h = 25$. We also evaluate model performance under different training sample sizes, as reported in Figure \ref{fig:sample_size}.

For this experiment, we adopt the same network architecture as in \cite{wehenkel2023robust}. Optimization is performed using AdamW \cite{loshchilov2017DecoupledWD, kingma2014AdamAM} with a learning rate of 0.001, a cosine annealing scheduler, and a weight decay of $5 \times 10^{-7}$. 

\paragraph{Reaction diffusion} 
In this experiment, we address a high-dimensional problem, where $\vect{x} \in \mathbb{R}^{2 \times 32 \times 32 \times \tau}$, governed by a two-dimensional reaction–diffusion PDE of the FitzHugh–Nagumo type:
\begin{equation}
\frac{du}{dt} = \underbrace{\Delta u + u - u^3 - v}_{f_p} - k, \quad \frac{dv}{dt} = \underbrace{b \Delta v + u - v}_{f_p}.
\end{equation}
Here, $\Delta$ denotes the Laplace operator, while $b$, and $k$ are system parameters, and $u(0), v(0)$ define the initial conditions. Trajectory samples are generated by drawing both the initial conditions and parameters from uniform distributions, following the setup of \cite{wehenkel2023robust}. The incomplete physics model $f_p$ omits the parameter $k$. Each simulation produces a state space representation $[u(t), v(t)] \in \mathbb{R}^{2 \times 32 \times 32}$.

We trained the models on three datasets containing $50, 25$ and $10$ trajectories, respectively. Each trajectory spans a time interval $[t_0, t_{\text{max}}] = [0.0, 1.0]$ and is divided into segments (history windows) of size $h = 5$. The test set consists of 250 trajectories with a longer time horizon of length 50 ($t_0 = 0.0, t_{\text{max}} = 5.0, \Delta t = 0.1$), used to evaluate extrapolation and generalization performance.

Our model adopts the same network architecture as in \cite{wehenkel2023robust}. Optimization is performed using AdamW with a learning rate of 0.001, a cosine annealing scheduler, and a weight decay of $5 \times 10^{-7}$.
\paragraph{Damped Pendulum} 
We generated the dataset by solving a non-linear damped pendulum:
\begin{equation}
\label{eq:pendulum-fric}
    \underbrace{\frac{\mathrm{d}^2 x(t)}{\mathrm{d} t^2} +\omega^2 \sin x(t)}_{f_p} + \underbrace{\xi\frac{dx(t)}{dt}}_{\text{missing physics}}=0,
\end{equation}
where the initial pendulum angle is $x(0) \sim \mathcal{U}(-1.57, 1.57)$, the angular frequency $\omega \sim \mathcal{U}(0.785, 3.14)$, and the damping coefficient $\xi \sim \mathcal{U}(0.6, 1.5)$. The incomplete physics model $f_p$ corresponds to a frictionless pendulum.
We train the model on 1000 training, 250 validation, and 100 test instances, each with trajectory length 200 ($t_0 = 0.0, t_{\text{max}} = 20, \Delta t = 0.1$), split into segment (history windows) of size $h = 25$.  

In this experiment, VGB-DM regresses 2D vector fields — velocity and acceleration — constructed via Lagrange interpolation using three consecutive points from each sampled trajectory segment, as described in \eqref{eq:second-order-dm}. The encoder architecture is identical to that of Phys-VAE. The learnt 2D vector field, $[v_t^\phi, a_t^\phi]$, is modelled using a shared MLP with two hidden layers of size [64, 64], followed by two distinct head networks (each with layers [64, 64]) corresponding to the velocity and acceleration components, respectively.  
We set the acceleration hyperparameter in \eqref{eq:second-order-dm} to $\alpha = 0.5$, which provides a balance between smoothness and curvature in the learnt dynamics. The model is optimized using AdamW with a learning rate of 0.001, a cosine annealing learning rate scheduler, and a weight decay of $5 \times 10^{-7}$.

\paragraph{Chaotic Lorenz Attractor System}
We analysed the Lorenz System for varying initial conditions and parameters. The system dynamics are governed by: 
\begin{align}
\frac{\mathrm{d}u}{\mathrm{d}t} &= \sigma (v - u), \\
\frac{\mathrm{d}v}{\mathrm{d}t} &= \underbrace{u (\rho - w)}_{\text{missing physics}} - v, \\
\frac{\mathrm{d}w}{\mathrm{d}t} &= u v - \beta w .
\end{align}

We generate the trajectory with $u(0), v(0), w(0)$ sampled from a Gaussian distribution and the parameters are also randomly sampled with $\sigma \sim \mathcal{U}(9.5, 10.5), \rho \sim \mathcal{U}(27.0, 29.0), \beta \sim \mathcal{U}(2.6, 2.8)$. Each trajectory
spans a time interval $t_0=0.0$ to $t_{\text{max}}=2.0$ and time resolution $\Delta t=0.0339$. The training dataset is made of $1000$ trajectories divided into segments of size $h = 30$. Instead, the validation and test set consists of $250$ trajectories of length 60. 

For our grey-box model, the known physics $f_p(\cdot ; \theta)$ incorporates the parameters $\theta = [\sigma, \beta]$. The term $u(\rho - w)$, representing buoyancy and non-linear advection, is treated as the unmodeled, stochastic part of the system. We remark that all latent parameters are inferred in a fully unsupervised manner in our method.

We designed a multi-layer gated recurrent unit (GRU) RNN-based encoder with 64 hidden states and two linear layers to infer both the physics parameters $\theta$ and the missing stochasticity $z = \rho$ directly from the observed trajectory. The velocity vector field $v^\phi$ is modelled using a simple conditional MLP with 4 hidden layers of 128 neurons. The model is optimized using AdamW with a learning rate of $0.001$, a cosine annealing learning rate scheduler, and a weight decay of $5 \times  10^{-7}$.

\subsection{Additional Results and Analyses}
\label{app:addition-results}

\begin{figure}[h]
    \centering
    \includegraphics[width=\linewidth]{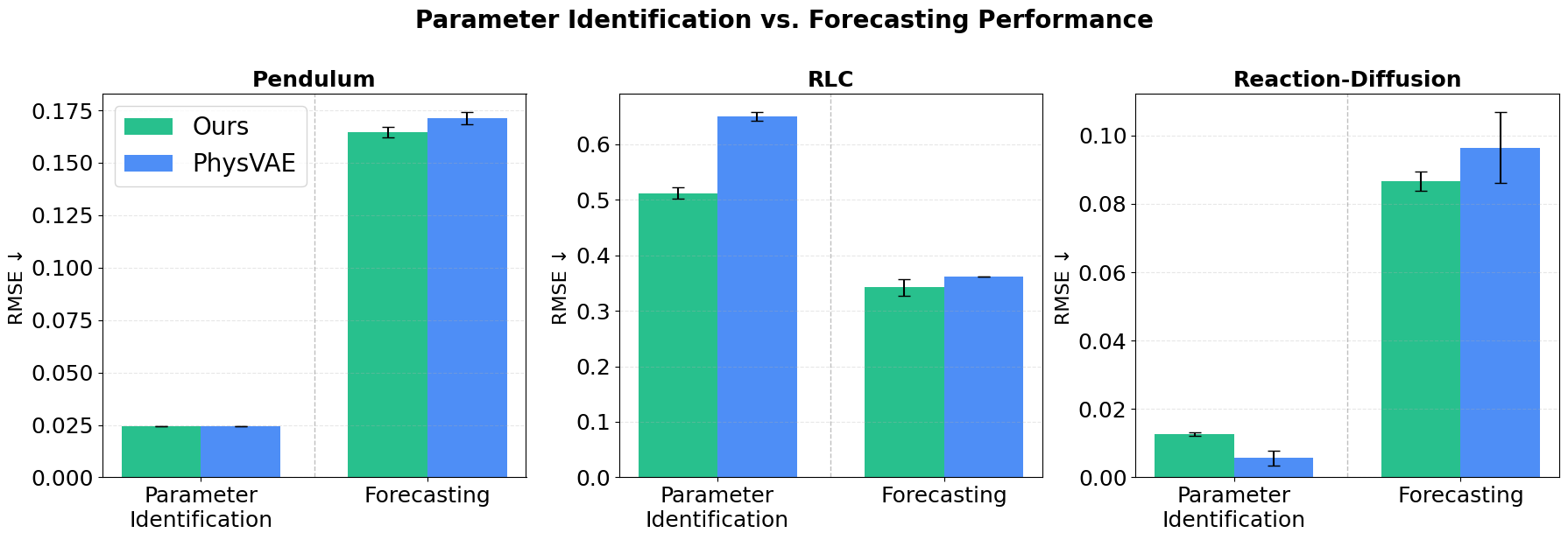}
    \caption{Parameter identification (inference) vs. forecasting performance across the three dynamical systems (lower RMSE is better).}
    \label{fig:params-identification}
\end{figure}

Figure~\ref{fig:params-identification} compares parameter identification (inference) and forecasting performance across the three dynamical systems. Our method achieves consistently better forecasting accuracy and comparable or slightly improved parameter inference. In particular, forecasting errors are lower across all systems, while parameter estimation remains on par with PhysVAE. Unlike PhysVAE, which requires careful regularization tuning for stable parameter recovery, our approach achieves balanced performance without such tuning, maintaining strong forecasting capability and robust generalization.

\paragraph{Analysis of Loss Decomposition}
The complete variational objective, $\mathcal{L}_{\text{VGB-DM}}^{\text{VI}}(\phi, \psi)$ is composed of three distinct terms: the Flow Matching (FM) loss which focuses on forecasting the dynamics , and the Physics Parameter KL-divergence (Ph-KL), and the Latent Variable KL-divergence (Z-KL), as defined in Equation~\ref{eq:GB-V-DynM}. To assess the relative contribution of each term to the total loss and to identify potential optimization bottlenecks, we report their mean values and standard deviations evaluated on the test set in Table~\ref{tab:results}. This decomposition allows for a clear comparison of which component dominates the overall objective for each experimental system.

\begin{align}
    \mathcal{L}_{\text{VGB-DM}}^{\text{VI}}(\phi, \psi) & = \mathbb{E}_{\substack{\pi(\vect{x}),\pi(\cdot | \vect{x})}} \Bigl[ \Bigr. \notag  \mathbb{E}_{\substack{q_\psi(z, \theta \mid \vect{x}_{k-h:k}) \\ t \sim \mathcal{U}(0,1)}} \underbrace{\Big\| (v_t^\phi(x_t \mid  \theta, z) \circ f_p(x_t, \theta)) 
    - \dot{x}_t \Big\|^2}_{\text{FM Loss}} \notag \\  
    & \quad - \underbrace{\text{KL}\Bigl[q_{\psi}(\theta \mid z, \vect{x}_{k-h:k}) \; \Vert \; p(\theta)\Bigr]}_{\text{Ph-KL}}  \\ 
    & \quad   -\underbrace{\text{KL}\Bigl[q_{\psi}(z \mid \vect{x}_{k-h:k}) \; \Vert  \; p(z) \Bigr]}_{\text{Z-KL}}
     \Bigl. \Bigr]
\end{align}

\begin{table}[h]
\centering
\caption{Evaluation of the N-ELBO terms}
\label{tab:results}
\begin{tabular}{lccc}
\toprule
\textbf{Experiment} & \textbf{FM} & \textbf{Ph-KL} & \textbf{Z-KL} \\
\midrule
Pendulum & $3.9 \times 10^{-5} \pm 1.8 \times 10^{-5}$ & $8.8 \times 10^{-4} \pm 1.1 \times 10^{-4}$ & $2.5 \pm 0.22$ \\
RLC      & $0.52 \pm 0.03$ & $3.4 \pm 1.4$ & $0.70 \pm 0.001$ \\
RD       & $3 \times 10^{-3} \pm 1 \times 10^{-3}$ & $2.7 \pm 1.1$ & $2 \times 10^{-3} \pm 2 \times 10^{-4}$ \\
Lorenz   & $3.8 \pm 1.5$ & $2.3 \pm 0.8$ & $0.003 \pm 0.001$ \\
\bottomrule
\end{tabular}
\end{table}

\subsubsection{Parameter Consistency Across Trajectory}
To validate the consistency of parameter estimates—given that our model infers them from input segment assumed to be representative—we evaluated their variation across each full trajectory. 

We computed the Coefficient of Variation (CV) for the estimated parameter values across the sliding windows of each trajectory. The CV, which normalizes the standard deviation by the mean, confirms the stability regardless of the parameter's scale. The median CV across all trajectories is 0.05 (or 5\%) or less, thus $\text{Median}\left(\frac{\sigma}{ |\mu|}\right) \le 0.05$. This demonstrates that the estimated physical parameters do not drift or fluctuate significantly during the episode, confirming the robustness of the inference method. We report their values for each experiment in Table \ref{table:params-consistency}.

\begin{table}[htb!]
\centering

\begin{tabular}{l c}
\hline
\textbf{Exp} & \textbf{Phys. Params. CV} \\
\hline
RLC & 0.0461, 0.0391 \\
Pendulum & 0.0496 \\
Reaction--Diffusion & 0.0072, 0.0031 \\
Lorenz & 0.0012, 0.0026\\
\hline
\end{tabular}
\caption{Median consistency of physics parameters across trajectory segments (test set).}
\label{table:params-consistency}
\end{table}

\section{Experimental details for Weather experiments} \label{app:details_weather_exp}

\subsection{Dataset details}
We utilize the preprocessed ERA5 dataset provided by WeatherBench \cite{rasp2020weatherbench}, a widely used benchmark framework for evaluating data-driven weather forecasting models. The original ERA5 data at $0.25^\circ$ resolution is regridded by WeatherBench to coarser resolutions of $5.625^\circ$, $2.8125^\circ$, and $1.40625^\circ$; in this work, we adopt the $5.625^\circ$ dataset at 6-hour intervals. Our study focuses on $K=5$ key variables: 2-metre temperature (t2m), atmospheric temperature (t), geopotential (z), and the 10-metre wind vector components (u10, v10). All variables are normalized to the range $[0,1]$ using min–max scaling. Among these, $z$ and $t$ are standard verification variables in medium-range Numerical Weather Prediction (NWP) models, while t2m and (u10, v10) are directly relevant to surface-level conditions that impact human activities. 

Following \cite{verma2024climode}, the dataset spans ten years of training data (2006–2015), one year for validation (2016), and two years for testing (2017–2018). The full set of ERA5 variables employed in our experiments is summarized in Table~\ref{tab:era5_vars}.

\begin{table}[h]
\centering
\caption{ECMWF data variables from ERA5 used in our dataset. \textit{Static} variables are time-independent, \textit{Single} represents surface-level variables, and \textit{Atmospheric} represents time-varying atmospheric properties at chosen altitudes.}
\label{tab:era5_vars}
\begin{tabular}{lllll}
\hline
Type & Variable name & Abbrev. & ECMWF ID & Levels \\
\hline
Static & Land-sea mask & lsm & 172 & \\
Single & 2 metre temperature & t2m & 167 & \\
Single & 10 metre U wind component & u10 & 165 & \\
Single & 10 metre V wind component & v10 & 166 & \\
Atmospheric & Geopotential & z & 129 & 500 \\
Atmospheric & Temperature & t & 130 & 850 \\
\hline
\end{tabular}
\end{table}

\subsection{Details of Climate VGB-DM}

\begin{figure}[ht!]
    \centering
    \includegraphics[width=\linewidth]{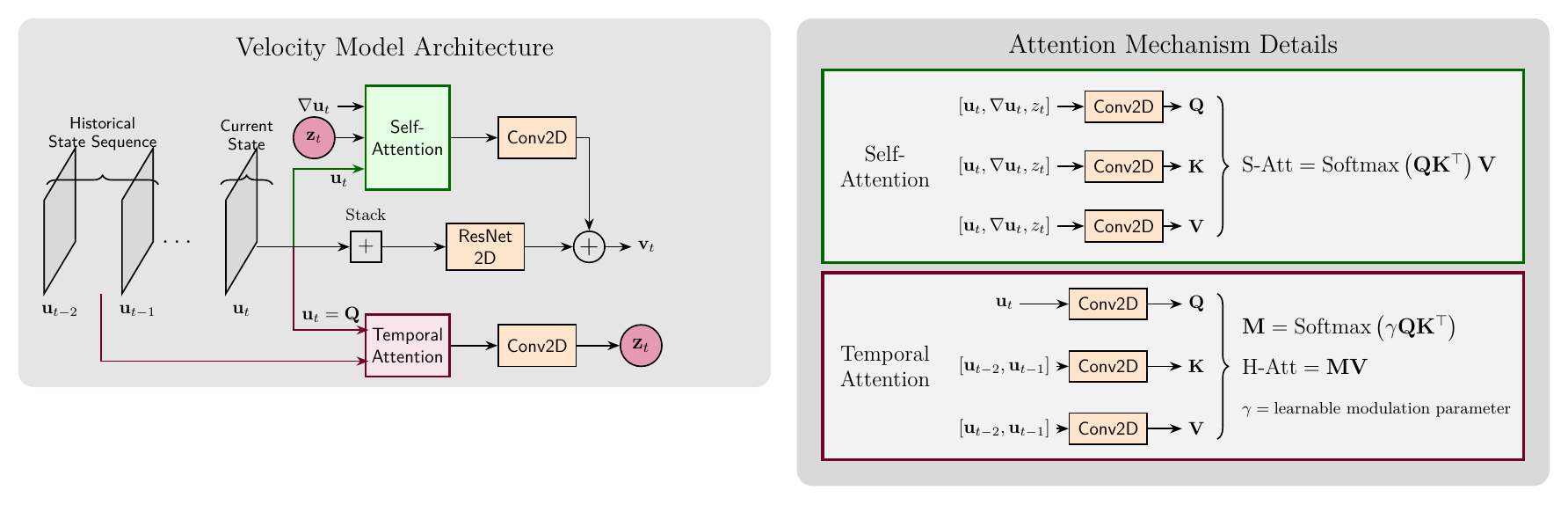}
    \caption{Diagram of the Velocity model architecture (left) and its lightweight attentions mechanism (right)}
    \label{fig:velocity-model-arch}
\end{figure}

For the climate modelling experiments in Section~\ref{sec:weather_dataset} of the paper, our neural architecture is based on the design presented in~\cite{verma2024climode}, adapted to our framework and excluding ClimODE’s external emission network $g$. Beyond the original ClimODE model, we additionally incorporate a lightweight temporal attention encoder applied to the trajectory states, with a window size $h=2$, and model the source terms.  A full description of our climate modelling framework is given in \eqref{eq:climate-modelling}, and the architecture of the velocity model is illustrated in the schematic of Figure \ref{fig:velocity-model-arch}. We optimize the model using AdamW with a learning rate of $5\times10^{-4}$.

The lightweight temporal attention encoder uses multi-head temporal attention in which the current state attends to a short sequence of past states independently at each spatial location. Queries are computed from the current state, while keys and values are obtained from the history via learned convolutional projections, preserving spatial resolution. Attention is applied exclusively over the temporal dimension, and the resulting history-aware features are projected back to the original channel space and aligned with the current state.
During training, the history consists of the previous two time steps, enabling the encoder to exploit true temporal context. At inference time, when only the current state is available, this state is replicated to match the required history length and used as the attention context.
This encoder processes the sampled state field $\vect{u}_{t-h:t}$ and produces a latent $z_t$, which is concatenated with the input to the grey-box model as described in the next paragraph\footnote{The implementation of the Climate VGB-DM model is provided in our \href{https://github.com/DMML-Geneva/VGB-DM}{VGB-DM GitHub repository}}. Although our main formulation allows for a variational posterior over the encoder latent variables, in practice we employ a deterministic temporal encoder, i.e. no KL regularization term is optimized. This choice is motivated by the fact that the underlying physics model does not involve latent physical parameters, but rather accounts for unmodelled dynamics such as velocity and source terms. Since all model evaluations are performed on forecasting performance, a deterministic encoder provides a simpler and effective alternative to a variational formulation without affecting the predictive objective.

\paragraph{Weather Grey-Box Dynamics.} For consistency with the notation introduced in~\cite{verma2024climode}, we remind the reader that spatio-temporal states are indicated by $u_t$, while $x$ is used for spatial coordinates. We outline the grey-box model of the advection type using a partial differential equation to represent the temporal evolution of the spatial field $u(x,t)$:
\begin{equation}
\label{eq:climate-modelling}
\frac{\partial u}{\partial t} = \underbrace{
-v_t^\phi(u_t, \nabla u_t, z_t, \psi) \cdot \nabla u_t 
 - u_t\,(\nabla \cdot v_t^\phi(u_t, \nabla u_t, z_t,  \psi)) + s_t^\phi(u_t, \nabla u_t, \psi)}_{V_t^\phi(u_t | z_t) = v_t^\phi(u_t | z_t) \circ f_p(u_t, v_t^\phi) \; \circ \; s^\phi_t(u_t, \nabla u_t, z_t,  \psi)},
\end{equation}
where $v_t^\phi$ is the learnable flow's velocity of the grey-box model and $s_t^\phi$ models the sources,  $z_t$ describes the latent variable from the encoder, finally the spatio-temporal embeddings is denoted by $\psi \in \mathbb{R}^{C \times H \times W}$, as in~\cite{verma2024climode} to keep the same notation. We also recall that $\nabla=\nabla_x$ denotes spatial gradients, and $\nabla \cdot v_t^\phi=\text{tr}(\nabla v_t^\phi)$ the divergence as denoted in ClimODE \cite{verma2024climode}.

\begin{figure*}[b!] 
\centering
\includegraphics[width=0.999\textwidth]{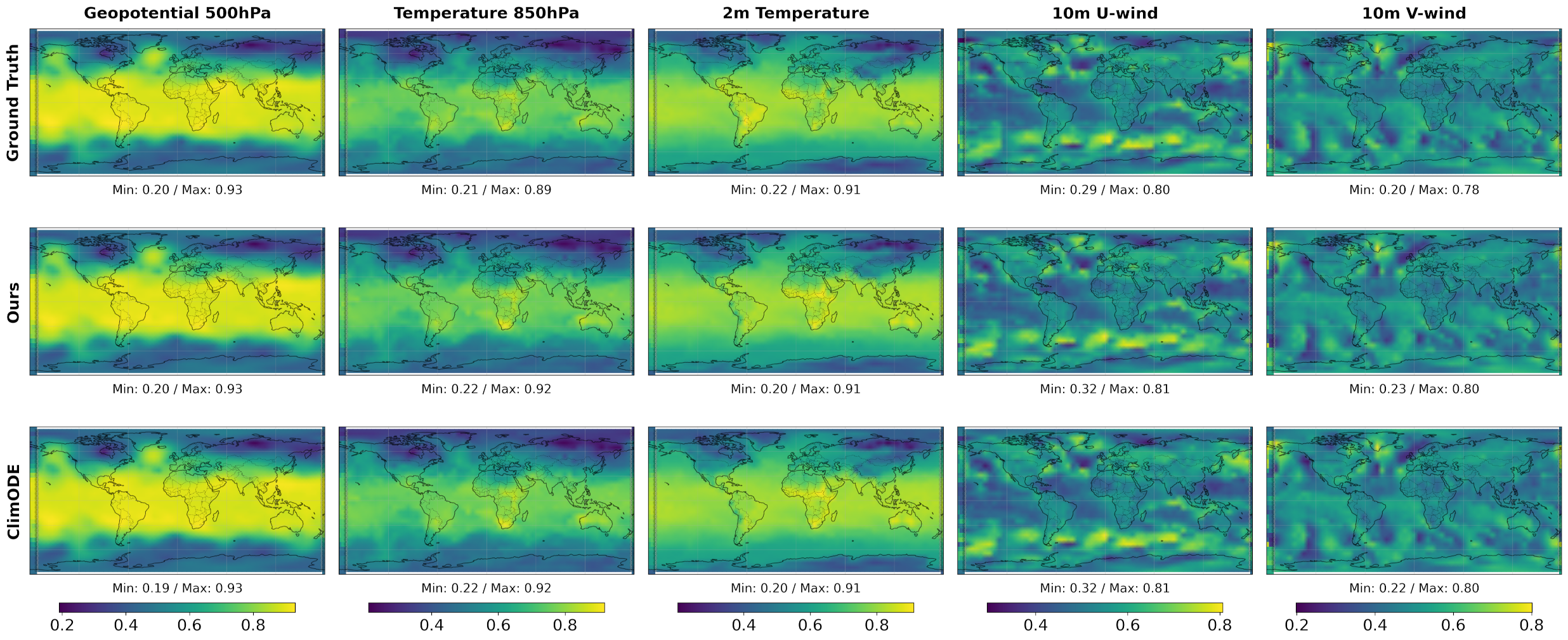}
\caption{Example of Climate forecasting using VGB-DM over $42$ hours.}
\label{fig:clime_predictions_hour}
\end{figure*}

\subsection{Time resolutions}
We have two resolutions, hourly over 42 hours, and monthly, over 5 months.
The hourly resolution results are in Figure~\ref{fig:clime_mse_global} and Figure~\ref{fig:residual_mse}.
The monthly resolution results are in Figure~\ref{fig:clime_mse_month}.

\subsection{Evaluation metrics}

Following~\cite{verma2024climode}, we assess benchmarks using latitude-weighted RMSE and Anomaly Correlation Coefficient (ACC) following the de-normalization of predictions.
$$
\text{RMSE} = \frac{1}{N}\sum_{t}\sqrt{\frac{1}{HW}\sum_{h}^{H}\sum_{w}^{W}\alpha(h)(y_{thw} - u_{thw})^{2}},\quad \text{ACC} = \frac{\sum_{t,h,w}\alpha(h)\tilde{y}_{thw}\tilde{u}_{thw}}{\sqrt{\sum_{t,h,w}\alpha(h)\tilde{y}_{thw}^{2}}\sqrt{\sum_{t,h,w}\alpha(h)u_{thw}^{2}}}
$$
where $\alpha(h) = \cos(h)/\frac{1}{H}\sum_{h}^{H}\cos(h')$ is the latitude weight and $\tilde{y} = y-C$ and $\tilde{u} = u - C$ are averaged against empirical mean $C = \frac{1}{N}\sum_{t}y_{thw}$.

\subsection{Predictions visualization}
In Figure~\ref{fig:clime_predictions_hour} we show the forecast of our model on the ERA5 dataset over 42 hours horizon.
In the main text we reported the error between true and predicted values to facilitate showing the difference between our model and ClimODE.

\section{Proofs and Derivations} \label{app:proofs}
In this section we prove that the marginal of the our velocity vector field are preserved, we demonstrate it by using the existing \textbf{Proposition 3.2 (Coupling Preservation)} of \cite{zhang2024tjfm}. In addition in the sub-section \ref{app:elbo-derivation}, we report the full ELBO derivation of our objective \ref{eq:GB-V-DynM}.

\subsection{Proof: Coupling Preservation with learnt Latents by Marginal Preservation}
\label{app:proof-marginal}
\paragraph{Setup}
Denote $\vect{x} := (x_{t_1}, x_{t_2}, \ldots, x_{t_T})$ as a trajectory sampled from the data distribution $\pi(\vect{x})$ and:
\begin{align}
p_t(x_t | x) &:= \mathcal{N}((t_{k+1} - t) x_{t_k} + (t - t_k) x_{t_{k+1}}, \sigma^2 (t_{k+1} - t) (t - t_k) \mathbf{I}), \label{eq:tfm:pt} \\
u_t(x_t | x) &:= \frac{x_{t_{k+1}} - x_t}{t_{k+1} - t}. \label{eq:tfm:flow}
\end{align}
where $p_t(x_t | x)$ is the Gaussian interpolant between consecutive trajectory points $x_{t_k}$ and $ x_{t_{k+1}}$ and $u_t(x_t | \vect{x})$ is the target velocity vector field.

Consider our simulation-free method, where the conditioning variable is given by latent representations produced by an encoder network. More formally, let $\mathbf{c} = (\theta, z) \in \mathbb{R}^{d_c}$ denote the concatenated latent variables obtained from an encoder $E_\psi: \mathcal{X}^T \to \mathbb{R}^{d_c}$ that maps trajectories to latent codes. The grey-box velocity field is defined as 
\[
V_t^{\phi}(x_t|\mathbf{c}) = v_t^\phi (x_t | \mathbf{c}) \circ f_p(x_t, \theta),
\]
where $V_t^{\phi}$ is the total velocity vector field incorporating the physics model, $u_t(x_t | \vect{x})$ is the true velocity field conditioned on the full trajectory $\vect{x}$, $q_\psi(\mathbf{c} | \vect{x})$ is the encoder posterior parameterised by $\psi$, and $p(\mathbf{c})$ is the prior distribution over the latents.

Our training objective is:
\begin{equation}
\mathcal{L}_{\text{VGB-DM}}(\phi,\psi) = 
\mathbb{E}_{t, \pi(\vect{x})} \left[ 
\mathbb{E}_{q_{\psi}(\mathbf{c}|\vect{x})} 
\left[ 
\| V_t^{\phi}(x_t|\mathbf{c}) - u_t(x_t|\vect{x}) \|^2 
\right] 
+ \mathrm{KL}(q_{\psi}(\mathbf{c}|\vect{x})\|p(\mathbf{c})) 
\right],
\end{equation}
where $V_t^{\phi}$ is the total velocity vector field incorporating the physics model, $u_t(x_t | \vect{x})$ is the true velocity field conditioned on the full trajectory $x$, $q_\psi(\mathbf{c} | \vect{x})$ is the encoder posterior parameterised by $\psi$, and $p(\mathbf{c})$ is the prior distribution over the latents.

\paragraph{Main Result} We prove that training with the learnt latent conditioning and the grey-box velocity field preserves the optimal coupling between trajectory distributions, ensuring the marginal vector field $u_t(x_t | \mathbf{c}) $ matches the true trajectory-conditioned $u_t(x_t | \vect{x})$ vector field in expectation.

Our proof schema follows as:

\begin{enumerate}
    \item We introduce \textbf{Assumption (A4)} characterizing the optimal encoder.
    \item We prove that \textbf{(A4)} ensures the marginal vector field and satisfies the conditions of \textbf{Proposition 3.2} (Coupling Preservation) \cite{zhang2024tjfm}.
    \item We recall \textbf{Proposition 3.2} to conclude that $\Pi(u)^\star = \Pi^\star(\vect{x}_{1:T})$.
\end{enumerate}

Throughout, we rely on the regularity conditions established in \textbf{Lemma A.1} of \cite{zhang2024tjfm} regarding Lipschitz continuity and integrability.

\paragraph{Assumption (A4): Optimal Posterior Encoder }\label{assumption:optimal_encoder}
The encoder learns the optimal posterior distribution, that is, there exists $\psi^*$ such that $q_{\psi^*}(\mathbf{c} | \vect{x}) = q^*(\mathbf{c} | \vect{x})$ where
\begin{equation}
q^*(\mathbf{c} | \vect{x}) = 
\arg\min_{q(\mathbf{c}|\vect{x})} 
\mathbb{E}_{t, x\sim \pi(\vect{x}), \mathbf{c} \sim q(\mathbf{c} | \vect{x}), x_t \sim p_t(x_t | \vect{x})} 
\left[ 
\| V_t^{\phi}(x_t|\mathbf{c}) - u_t(x_t | \vect{x}) \|_2^2 
+ \mathrm{KL}(q(\mathbf{c}|\vect{x})\|p(\mathbf{c})) 
\right] = 0    
\end{equation}

This assumption says that the encoder captures sufficient trajectory-specific information in the latent code $\mathbf{c} = (\theta, z)$ such that conditioning on $\mathbf{c}$ allows perfect recovery of the trajectory-dependent vector field $u_t(x_t | \vect{x})$ via the grey-box model $V_t^{\phi}$ (i.e., the minimal MSE is zero), while the KL divergence encourages the posterior to match the prior.


\subsection*{Main Proposition}

\begin{proposition}[\textbf{Coupling Preservation with learnt Latents by Marginal Preservation}]
\label{prop:tfm_latent}
Under the regularity conditions of Lemma A.1 \cite{zhang2024tjfm}, if \textbf{Assumption (A4)}  holds and $x, \pi(\vect{x}), p_t(x_t|\vect{x}),$ and $u_t(x_t|\vect{x})$, then training with $\mathcal{L}_{\text{VGB-DM}}(\phi, \psi^*)$ preserves the optimal coupling:
\[
\Pi(u)^\star = \Pi^\star(\vect{x}_{1:T}).
\]
\end{proposition}
\subsection{Proof of Proposition \ref{prop:tfm_latent}}

\paragraph{Step 1: Recall Proposition 3.2.}\cite{zhang2024tjfm} establishes that under mild regularity conditions, if
\begin{equation}
    \mathbb{E}_{t, x\sim \pi(\vect{x}), \mathbf{c} \sim q(\mathbf{c} | \vect{x}), x_t \sim p_t(x_t | \vect{x})} 
\| u_t(x_t | \vect{x}, \mathbf{c}) - u_t(x_t | \mathbf{c}) \|_2^2 = 0,
\end{equation}
then the push-forward flows satisfy $\int_0^T u_t(x_t |\vect{x}, \mathbf{c}) dt = \phi(x_0, \mathbf{c}, \vect{x}) = \phi(x_0, \mathbf{c})=\int_0^T u_t(x_t | \mathbf{c}) dt$ and consequently $\Pi(u)^\star = \Pi^\star(\vect{x}_{1:T})$.

We aim to confirm this condition by applying \textbf{Assumption (A4)} (see \textbf{Steps 2-3}), observing that the grey-box speed $V_t^{\phi}$ serves as an approximation for $u_t$ under optimal conditions.
\paragraph{Step 2: Marginal Vector Field.}
The marginal vector field conditioned on $\mathbf{c}$ is critical because it determines the push-forward flow $\phi(x_0, \mathbf{c}) = \int_0^T u_t(x_t | \mathbf{c}) dt$ that defines the coupling $\Pi(u)$.  Therefore, by marginalising it over trajectories, we have:
\begin{equation*}
u_t(x_t | \mathbf{c}) = \mathbb{E}_{\vect{x} \sim \pi(\vect{x} | \mathbf{c})} [u_t(x_t | \vect{x})].    
\end{equation*}
By Bayes' rule, the posterior over trajectories given the latent code is:
\begin{equation*}
    \pi(\vect{x} | \mathbf{c}) = \frac{q(\mathbf{c} | \vect{x}) \pi(\vect{x})}{\int q(\mathbf{c} | \vect{x}') \pi(\vect{x}') d\vect{x}'}.
\end{equation*}
When the encoder satisfies \textbf{Assumption (A4)}, we have $q(\mathbf{c} | \vect{x}) = q^*(\mathbf{c} | \vect{x})$, and therefore:
\begin{equation}
u_t(x_t | \mathbf{c}) = \mathbb{E}_{\vect{x} \sim \pi^*(\vect{x} | \mathbf{c})} [u_t(x_t | \vect{x})].    
\end{equation}
At optimality, $V_t^{\phi}(x_t | \mathbf{c})$ matches this marginal.

\paragraph{Step 3: Verification of Proposition 3.2 Condition.}
We now show that Assumption (A4) implies the condition required by Proposition 3.2. Consider:
\begin{align*}
&\mathbb{E}_{t, x\sim \pi(\vect{x}), \mathbf{c} \sim q^*(\mathbf{c} | \vect{x}), x_t \sim p_t(x_t | \vect{x})} 
\| u_t(x_t | \vect{x}, \mathbf{c}) - u_t(x_t | \mathbf{c}) \|_2^2 \\
&= \mathbb{E}_{t, x\sim \pi(\vect{x}), \mathbf{c} \sim q^*(\mathbf{c} | \vect{x}), x_t \sim p_t(x_t | \vect{x})} 
\| u_t(x_t | \vect{x}) - u_t(x_t | \mathbf{c}) \|_2^2,
\end{align*}
where the equality follows since $u_t(x_t | x, \mathbf{c}) = u_t(x_t | \vect{x})$ (the true vector field depends only on the trajectory $x$).

By the Tower property of conditional expectation:
\begin{align*}
&\mathbb{E}_{t, x\sim \pi(\vect{x}), \mathbf{c} \sim q^*(\mathbf{c} | \vect{x}), x_t \sim p_t(x_t | \vect{x})} 
\| u_t(x_t | \vect{x}) - u_t(x_t | \mathbf{c}) \|_2^2 \\
&= \mathbb{E}_{t, \mathbf{c}} 
\left[ 
\mathbb{E}_{x \sim q^*(x | \mathbf{c}), x_t \sim p_t(x_t | \vect{x})} 
\left[ 
\| u_t(x_t | \vect{x}) - u_t(x_t | \mathbf{c}) \|_2^2 
\,\Big|\, \mathbf{c} 
\right] 
\right].
\end{align*}

For the optimal encoder $q^*(\mathbf{c} | \vect{x})$, by its defining property in Assumption (A4), we have:
\begin{equation}
u_t(x_t | \mathbf{c}) = \mathbb{E}_{x \sim q^*(x | \mathbf{c})} [u_t(x_t | \vect{x})],    
\end{equation}
which is the best predictor of $u_t(x_t | \vect{x})$ given $\mathbf{c}$ in terms of $L^2$ norm. Moreover, since Assumption (A4) specifies that the minimal value of the MSE objective is zero (with the KL term providing regularization but not affecting the zero-MSE at optimality), the conditional variance is zero:
\[
\mathbb{E}_{x \sim q^*(x | \mathbf{c}), x_t \sim p_t(x_t | \vect{x})} 
\left[ 
\| u_t(x_t | \vect{x}) - u_t(x_t | \mathbf{c}) \|_2^2 
\,\Big|\, \mathbf{c} 
\right] = 0
\]
for each fixed $\mathbf{c}$ and $t$. Therefore:
\begin{equation}
\mathbb{E}_{t, x\sim \pi(\vect{x}), \mathbf{c} \sim q^*(\mathbf{c} | \vect{x}), x_t \sim p_t(x_t | \vect{x})} 
\| u_t(x_t | \vect{x}) - u_t(x_t | \mathbf{c}) \|_2^2 = 0.    
\end{equation}

\paragraph{Step 4: Application of Proposition 3.2.}
Since the condition of Proposition 3.2 is satisfied under Assumption (A4), and all regularity conditions from Lemma A.1 hold (Lipschitz continuity of $\delta_{\text{data}}$ and appropriate integrability conditions for exchange of integrals), we conclude that the push-forward flows satisfy:
\begin{equation}
\phi(x_0, \mathbf{c}, \vect{x}) = \phi(x_0, \mathbf{c})    
\end{equation}
for all $x_0$. Therefore, the optimal coupling is preserved:
\begin{equation}
\Pi(u)^\star = \Pi^\star(\vect{x}_{1:T}).    
\end{equation}
\hfill $\square$

\paragraph{Some Practical Considerations}: \textbf{Assumption (A4)} can be approximately satisfied through the following training strategy:
\begin{itemize}
    \item \textbf{Joint optimization:} Train both the encoder $E_\psi$ and the velocity field components ($v_\phi$ and the physics-informed $f_p$) to jointly minimize $\mathcal{L}_{\text{VGB-DM}}(\phi, \psi)$.
    \item \textbf{Adequate capacity:} Make sure the latent dimension $d_c = \dim((\theta, z))$ is large enough to encapsulate information specific to the trajectory (it acts as a sufficient statistic for trajectory), encompassing the parameters of the physics model.
\end{itemize}

\subsection{ELBO Derivation}
\label{app:elbo-derivation}
\begin{align}
\log p_t^\phi(\dot{x}_t \mid x_k) 
&= \log \int p_t^\phi(\dot{x}_t, \theta, z \mid x_k)\, d\theta\, dz \\
&= \log \int q_{\psi}(z, \theta \mid \vect{x}) \frac{p_t^\phi(\dot{x}_t, \theta, z \mid x_k)}{q_{\psi}(z, \theta \mid \vect{x})}\, d\theta\, dz \\
&\geq \int q_{\psi}(z, \theta \mid \vect{x}) \log \frac{p_t^\phi(\dot{x}_t, \theta, z \mid x_k)}{q_{\psi}(z, \theta \mid \vect{x})}\, d\theta\, dz \quad \text{(Jensen's inequality)} \\
&= \int q_{\psi}(z, \theta \mid \vect{x}) \log \frac{p_t^\phi(\dot{x}_t \mid \theta, z, x_k) p_t^\phi( \theta, z \mid x_k)}{q_{\psi}(z, \theta \mid \vect{x})}\, d\theta\, dz\\
&= \mathbb{E}_{q_{\psi}(z, \theta \mid \vect{x})}\left[\log p_t^\phi(\dot{x}_t, \theta, z \mid x_k) - \log \frac{q_{\psi}(z, \theta \mid \vect{x})}{p_t^\phi( \theta, z \mid x_k)}\right] \\
&= \mathbb{E}_{q_{\psi}(z, \theta \mid \vect{x})}[\log p_t^\phi(\dot{x}_t \mid \theta, z, x_k)] - \text{KL}\left[q_{\psi}(z, \theta \mid \vect{x}) \Vert p_t^\phi(\theta, z \mid x_k)\right]
\end{align}

Assuming independent and data and time independent priors:
\begin{equation}
   p_t^\phi(\theta, z \mid x_k) = p(\theta )p(z) 
\end{equation}
Assuming joint conditional \textbf{dependency} of:

    \begin{equation}
        q_\psi(z, \theta \mid \vect{x}) = q_\psi(\theta \mid  z,  \vect{x}) q_\psi(z \mid \vect{x})
    \end{equation}
    The objective becomes:
    \begin{align}
        \log p_t^\phi(\dot{x}_t \mid x_k) &\geq \\ \nonumber
        & \mathbb{E}_{q_{\psi}(z, \theta \mid \vect{x})}[\log p_t^\phi(\dot{x}_t \mid \theta, z, x_k)] 
        - \mathbb{E}_{q_{\psi}(z \mid \vect{x})}\text{KL}\left[q_{\psi}(\theta \mid z, \vect{x}) \Vert p(\theta)\right] 
        -  \text{KL}\left[q_{\psi}(z \mid \vect{x}) \Vert p(z)\right]
    \end{align}

\subsection{Reduced Complexity and Optimization Stability}
\label{app:reduced_complexity}

We provide an additional intuition for the empirical sample efficiency and optimization stability observed in our method. These improvements arise from two complementary mechanisms: a reduction in functional complexity induced by the incorporation of an incomplete physics model and a more favourable optimisation landscape enabled by a simulation-free training objective.

\paragraph{Reduced Functional Complexity.}
Compared to fully black-box models such as BB-NODE or VBB-DM, our approach benefits from a strong inductive bias introduced through the incomplete physics model $f_p(x,\theta)$. In black-box formulations, the learning algorithm must explore a large function space to recover the full system dynamics from data alone. In contrast, by explicitly integrating $f_p$, we impose a structural constraint that encodes known physical principles. As a result, the learning problem is simplified: rather than discovering physical laws from scratch, the model focuses on inferring a low-dimensional set of unknown parameters $\theta$ and latent variables $z$, while learning a residual or corrective dynamics term $v_\phi$. This reduction in hypothesis space directly contributes to improved sample efficiency.

\paragraph{Optimization Landscape and Simulation-Free Training.}
While physics-informed latent variable models such as PhysVAE also benefit from reduced functional complexity, our method exhibits faster convergence and improved empirical sample efficiency. This difference can be attributed to the choice of training objective. PhysVAE relies on a simulation-based loss that requires integrating the system dynamics through a numerical ODE solver over a time horizon $T$, of the form
\[
\mathcal{L}_{\text{sim}} \approx 
\left\| X - \mathrm{ODESolve}(v^\phi \circ f_p, x_0, T) \right\|^2 .
\]
The gradient $\nabla \mathcal{L}_{\text{sim}}$ must be computed by backpropagating through the numerical solver, resulting in a computational cost of $\mathcal{O}(L)$, where $L$ denotes the number of function evaluations performed by the solver.

In contrast, our method optimizes a simulation-free dynamics matching objective, which directly aligns the learned vector field with a target velocity $\dot{x}_t$ obtained from data interpolation:
\[
\mathcal{L}_{\text{DM}} \approx 
\mathbb{E}_t \left[ \left\| (v^\phi_t \circ f_p) - \dot{x}_t \right\|^2 \right].
\]
The gradient $\nabla \mathcal{L}_{\text{DM}}$ is computed via standard backpropagation and requires only a single forward pass through the network, yielding $\mathcal{O}(1)$ computational complexity per update.

\paragraph{Lemma 1 - Gradient Variance and Convergence. }
Beyond computational cost, the simulation-free formulation leads to more stable optimization dynamics. Estimating $\nabla \mathcal{L}_{\text{sim}}$ is known to be challenging due to the need to differentiate through a numerical ODE solver. This process can introduce high gradient variance, susceptibility to exploding or vanishing gradients—particularly for long integration horizons or stiff systems—and the accumulation of numerical errors. In contrast, $\nabla \mathcal{L}_{\text{DM}}$ corresponds to a supervised regression gradient, which avoids integration-induced instabilities.

Formally, this suggests that the variance of the gradient estimator under the simulation-free objective is lower than that of the simulation-based objective,
\[
\operatorname{Var}(\nabla \mathcal{L}_{\text{DM}}) 
< 
\operatorname{Var}(\nabla \mathcal{L}_{\text{sim}}),
\]
which in turn implies faster convergence rates under stochastic optimization. This reduction in gradient variance, combined with the reduced functional complexity discussed above, explains the observed improvements in both optimization stability and sample efficiency.

\end{document}